\documentclass{article} 
\usepackage{iclr2021_conference,times}

\usepackage{amsmath,amsfonts,bm}

\def\eqref#1{equation~\ref{#1}}

\def\1{\bm{1}}

\DeclareMathAlphabet{\mathsfit}{\encodingdefault}{\sfdefault}{m}{sl}
\SetMathAlphabet{\mathsfit}{bold}{\encodingdefault}{\sfdefault}{bx}{n}

\newcommand{\R}{\mathbb{R}}

\usepackage[utf8]{inputenc} 
\usepackage[T1]{fontenc}    
\usepackage{hyperref}       
\usepackage{url}            
\usepackage{booktabs}       
\usepackage{amsfonts}       
\usepackage{nicefrac}       
\usepackage{microtype}      

\usepackage{float}
\usepackage{graphicx}

\usepackage{color}
\usepackage{amsmath}
\usepackage{amssymb}
\usepackage{amsthm}
\usepackage{mathtools}
\usepackage[mathscr]{eucal}
\usepackage{bm}
\usepackage{bbm}
\usepackage{relsize}
\usepackage{graphics}
\usepackage{wrapfig}
\usepackage{multirow}
\usepackage{enumitem}
\usepackage{tablefootnote}

\usepackage{array}

\usepackage{pgf}
\usepackage{xinttools}
\usepackage{tikz}
\usetikzlibrary{arrows.meta}
\usepackage{textcomp}
\usetikzlibrary{calc,shapes,arrows,positioning,automata,trees}
\usepackage{placeins} 
\usepackage{tabularx}
\usepackage{graphicx}
\usepackage{subcaption} 
\usepackage{listings}
\usepackage{xargs,lipsum,caption,changepage,ifthen}

\usepackage{algorithm}
\usepackage{algpseudocode}

\usepackage[toc,page]{appendix}

\newtheorem*{theorem*}{Theorem}
\newtheorem*{definition*}{Definition}

\newcommand\Bn{\bm{n}}

\newcommand\Bp{\bm{p}}

\newcommand\Bv{\bm{v}}

\newcommand\Bx{\bm{x}}

\newcommand\BB{\bm{B}}

\newcommand\BE{\bm{E}}

\newcommand\BP{\bm{P}}
\newcommand\BQ{\bm{Q}}

\newcommand\BT{\bm{T}}

\newcommand\BZero{\bm{0}}
\newcommand\sgn{\mathop{\mathrm{sgn}\,}}

\newcommand{\PAR}[1]{{{\left(#1\right)}}} 

\renewcommand{\leq}{\leqslant}

\usepackage{pdflscape} 

\newcolumntype{R}[1]{>{\raggedright\arraybackslash}p{#1}}
\newcolumntype{C}[1]{>{\centering\arraybackslash}p{#1}}
\newcolumntype{L}[1]{>{\raggedleft\arraybackslash}p{#1}}

\usepackage{bigdelim}
\usepackage{colortbl} 
\definecolor{mColor1}{rgb}{0.95,0.95,0.95}

\usepackage{tikz-imagelabels}
\usepackage{calc}
\usepackage{gensymb}

\usepackage{overpic}
\usepackage{tikz}

\title{Learning 3D Granular Flow Simulations}

\author{\vspace{0.1cm}
Andreas Mayr\footnotemark[1] \quad
Sebastian Lehner\footnotemark[1] \quad
Arno Mayrhofer\footnotemark[2] \quad 
Christoph Kloss\footnotemark[2] \quad  \\ \vspace{0.1cm} \bf
Sepp Hochreiter\footnotemark[1]~$~^{,}$\footnotemark[3] \quad 
Johannes Brandstetter\footnotemark[1]~$~^{,}$\footnotemark[4]  \quad\\
\footnotemark[1]~~ELLIS Unit Linz, LIT AI Lab, Institute for Machine Learning,\\
                ~~Johannes Kepler University Linz, Austria\\
\footnotemark[2]~~DCS Computing GmbH, Linz, Austria\\ 
\footnotemark[3]~~Institute of Advanced Research in Artificial Intelligence (IARAI)\\
\footnotemark[4]~~University of Amsterdam, Amsterdam, Netherlands\\ 
\\
}

\iclrfinalcopy 
\begin{document}

\maketitle

\begin{abstract}
Recently, the application of machine learning models has gained momentum in natural sciences and engineering, which is a natural fit due to the abundance of data in these fields.
However, the modeling of physical processes from simulation data 
without first principle solutions remains difficult.
Here, we present a Graph Neural Networks approach towards accurate modeling 
of complex 3D granular flow simulation processes created by the discrete element method LIGGGHTS and concentrate on simulations of physical systems found in real world applications like rotating drums and hoppers.
We discuss how to implement Graph Neural Networks that deal with 3D objects, boundary conditions, particle - particle, and particle - boundary interactions such that an accurate modeling of relevant physical quantities is made possible. Finally, we compare the machine learning based trajectories to LIGGGHTS trajectories in terms of particle flows and mixing entropies.

\end{abstract}

\section{Introduction}
Granular flows are ubiquitous in nature and in a large array of industrial processes.
Pharmaceutical powders, plastic granulate and rocks obtained by mining are just some examples of granular media that are used in industries and which are processed in a multitude of different flow states.
While attempts have been made to formulate governing equations for granular flows, like the Navier-Stokes equations for fluid flow, they have so far eluded a general framework \citep{faccanoni2013}.
More recently, research focus has been expanded 
towards applications of Deep Learning in the simulation of  
physical domains, such as fluid dynamics, deformable materials, or aerodynamics~\citep{SanchezGonzalez2020, Pfaff2020}. 
One key component of the recent progress in deep neural network simulation is the usage of 
graph neural networks~\citep[][GNNs]{Scarselli2008}.
GNNs have demonstrated effectiveness in settings that involve interactions between many entities via forward dynamics~\citep{Battaglia2018}.
In~\citet{SanchezGonzalez2020}, GNNs
are used to obtain models that generalize fluid dynamics and 2D interactions over many timesteps and different initial conditions.
In~\citet{Pfaff2020}, mesh-based simulations 
are learned using GNNs to predict the dynamics of a wide range of physical
systems, including aerodynamics, or structural mechanics.

Compared to previous work \citep[i.e., ][]{SanchezGonzalez2020, Pfaff2020}, we focus on learning 3D simulations of granular particle flow with nontrivial geometric boundary conditions. These simulations are highly relevant for the design of industrial processes and allow to understand and improve particle flow dynamics of various given materials.
Since no underlying governing equation for general granular flows exists, we use simulation data originating from the Discrete Element Method \citep[][DEM]{Cundall1979} as the ground truth (see Appendix \ref{DEM} for further details on DEM).
We generate granular flow simulation data with the open-source DEM software LIGGGHTS \citep{Kloss2012}. LIGGGHTS allows the simulation of particulate flows with a wide range of materials and complex mesh-based wall geometries, and therefore enables the simulation of relevant industrial processes. Such complex mesh-based wall geometries are in contrast to \citet{SanchezGonzalez2020}, where for the 3D simulations static cuboids have been assumed and a material point method  \citep[][MPM]{sulsky1995application} based simulator \citep{Hu2018} has been applied.

Similar to~\citet{SanchezGonzalez2020}, we learn a time-transition model, consisting of an encoder, processor, and, a decoder,  to predict particle accelerations and build upon their relative positional encoding variant.
The time transition model from time $t_k$ to time $t_{k+1}$
is given by
\begin{align}
\dot{\Bp}^{t_{k+1}}=\dot{\Bp}^{t_{k}}+\Delta t\; \ddot{\Bp}^{t_{k+1}} \ , \label{ttmodel} \\
\Bp^{t_{k+1}}=\Bp^{t_{k}}+\Delta t\; \dot{\Bp}^{t_{k+1}} \ ,  
\end{align}
where $\Bp$ is the particle location, and $\dot{\Bp}$ the 
particle velocity. The new particle acceleration $\ddot{\Bp}$ at time $t_{k+1}$ needs to be predicted. 
The contributions of this work towards an accurate modeling of complex 3D granular flows are three-fold:
\begin{itemize}
    \item Our model deals with triangularization of 3D objects, which is a very common representation in engineering applications. We achieve this by inserting virtual nodes if particles are close to boundaries and computing the proximity to the closest triangle. 
    \item We include particle - boundary interactions by forcing the network to be invariant with respect to the orientation of the normal vectors.
    \item We compare and analyze relevant physical quantities between simulated processes and the output of our network.
\end{itemize}

\section{Accurate Modeling of Granular Flow Dynamics}\label{sec:GranuFlow}

\textbf{Learning simulations that are governed by complex geometries. } 
Triangularization of geometric boundaries is very common in engineering applications.
One idea to tackle this challenge for 2D scenes is the  insertion of stationary, virtual particles into the scene to describe boundaries~\citep{SanchezGonzalez2020}. However, for 3D scenes and thus for many practically relevant engineering applications, 
a computationally more efficient approach is required for the following reasons:
Firstly, triangle surface areas in 3D would require significantly more stationary particles for the representation than curves in 2D scenes. Secondly, for certain time frames, only some parts of the mesh are relevant, e.g. as long as particles are in the state of a free fall in a container and the bottom of a container is still far away, the mesh part describing the bottom of the container is irrelevant for the next time step. 

\begin{figure}
\begin{center}
\includegraphics[trim=0 10 0 0, scale=0.4]{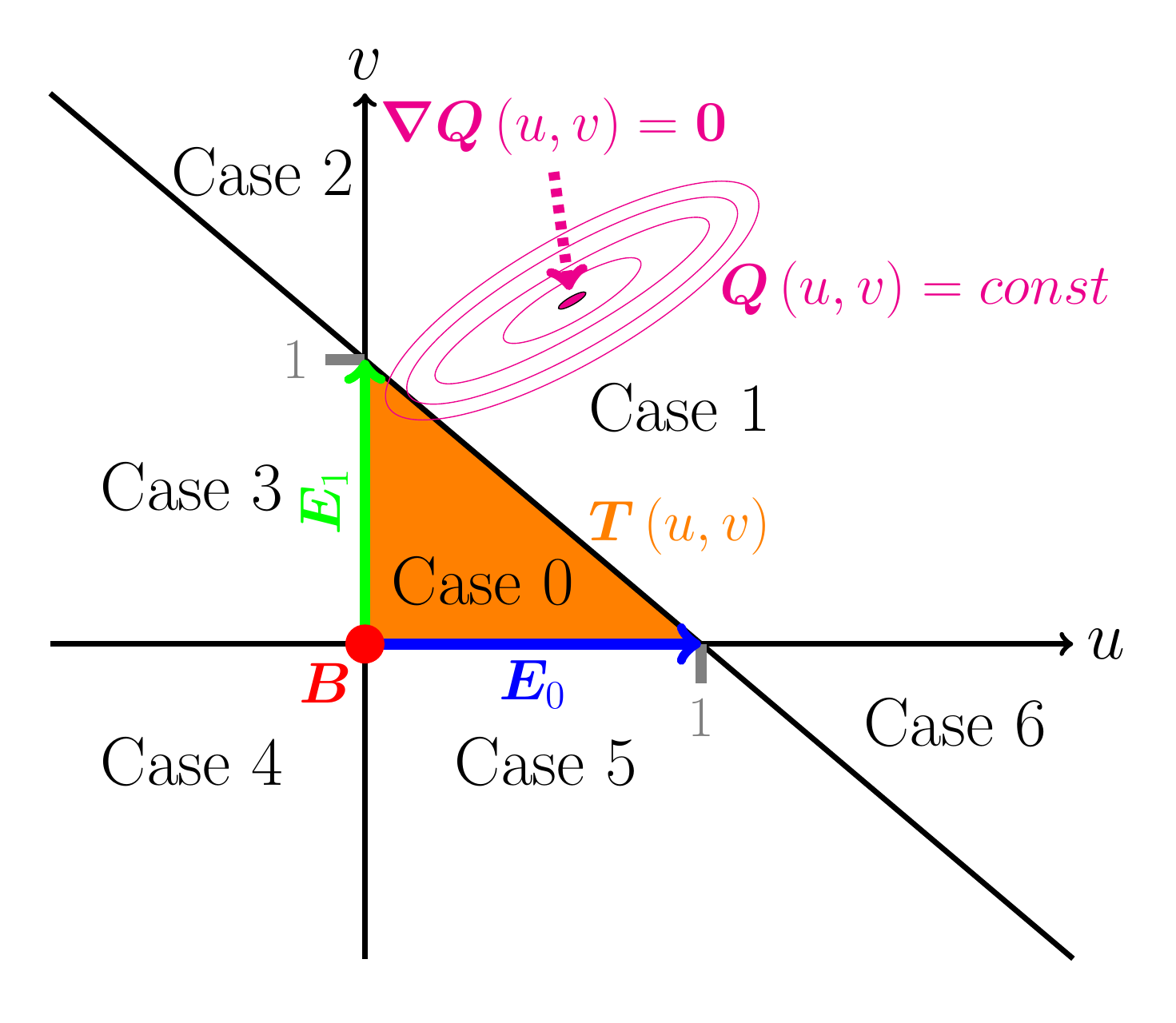}
\end{center}
\caption{Visualization of point - triangle distance calculations in 3D. The triangle is represented by a parameterized function $\BT\left(u,v\right)=\BB+u\;\BE_0+v\;\BE_1$ with $u\geq0$, $v\geq0$, $u+v\leq1$ (indicated by the orange area). Level sets of $\BQ\left(u,v\right)$ are indicated by ellipses and describe the squared Euclidean distance of a triangle point $\BT\left(u,v\right)$ to the point $\BP$, for which we compute the minimum distance.} 
\label{myTriangle}
\end{figure}

\newcommand\BNabla{\bm{\nabla}}
\textbf{Modeling distances to triangular boundaries in 3D scenes. } 
 The challenge is to correctly incorporate the modeling of the proximity of a particle to a triangular 3D boundary and at the same time avoid large computational cost.
Firstly, to avoid large computational cost, we dynamically insert boundary particles if a real particle is close to the corresponding boundary.
The inserted virtual nodes in the graph describe whole surface areas in contrast to single 3D points. Therefore, we indicate the particle type, virtual or real, by introducing additional node features, such that the neural network is able to distinguish them. In other terms, this allows the model to learn different dynamics for particle - particle and particle - boundary interactions.
The additional node features are: (i) type feature, i.e.,  a binary indicator of whether a node represents a particle that is real or virtual, and, in the latter case, (ii) the components of the normal vector of the triangular surface (null vectors for real particles).
The second challenge is to correctly model the proximity of a 
particle to the boundary, which
is computed as the minimum distance between the particle center and the closest point of the mesh triangles (adopted from \citet{Eberly1999}). 
We assume a triangle to be represented by a function $\BT$, which is parameterized by two scalar values $u$ and $v \in \R$:
\begin{equation}
    \BT\left(u,v\right)=\BB+u\;\BE_0+v\;\BE_1,
\end{equation}
where $u\geq0$, $v\geq0$, $u+v\leq1$, $\BB$ represents one of the corner points of the triangle, and, $\BE_0$ and $\BE_1$ are vectors from $\BB$ towards the other two corner points (see Figure~\ref{myTriangle}). We solve the optimization problem
\begin{align}
& \min_{u,v}\; \BQ\left(u,v\right)=\left|\BT\left(u,v\right)-\BP\right|^2 \\
s.t.\;\; & u\geq0,\;v\geq0,\;u+v\leq1 \nonumber
\end{align}
to obtain the minimizing parameters $u,v$ for retrieving the closest point $\BT\left(u,v\right)$ of the triangle from the point $\BP$ in an Euclidean sense. As indicated in Figure~\ref{myTriangle}, we have to distinguish seven cases: the simplest one is \textit{Case 0}, where $\BNabla \BQ\left(u,v\right)=\BZero$ is fulfilled in the inner of the triangle (i.e., $u>0$, $v>0$, $u+v<1$). If $\BNabla \BQ\left(u,v\right)=\BZero$ is fulfilled outside from that triangle, cases where the minimizing point from the triangle is at the triangle border, i.e., the lines $u=0, v=0, u+v=1$ have to be considered. In order to find the minimizing parameters in this case we have to look for stationary points of the one-parameter functions $\BQ\left(u,1-u\right)$, $\BQ\left(u,0\right)$ and  $\BQ\left(0,u\right)$.

\begin{figure}[ht]
\begin{center}
\includegraphics[trim=130 0 0 0, scale=0.27]{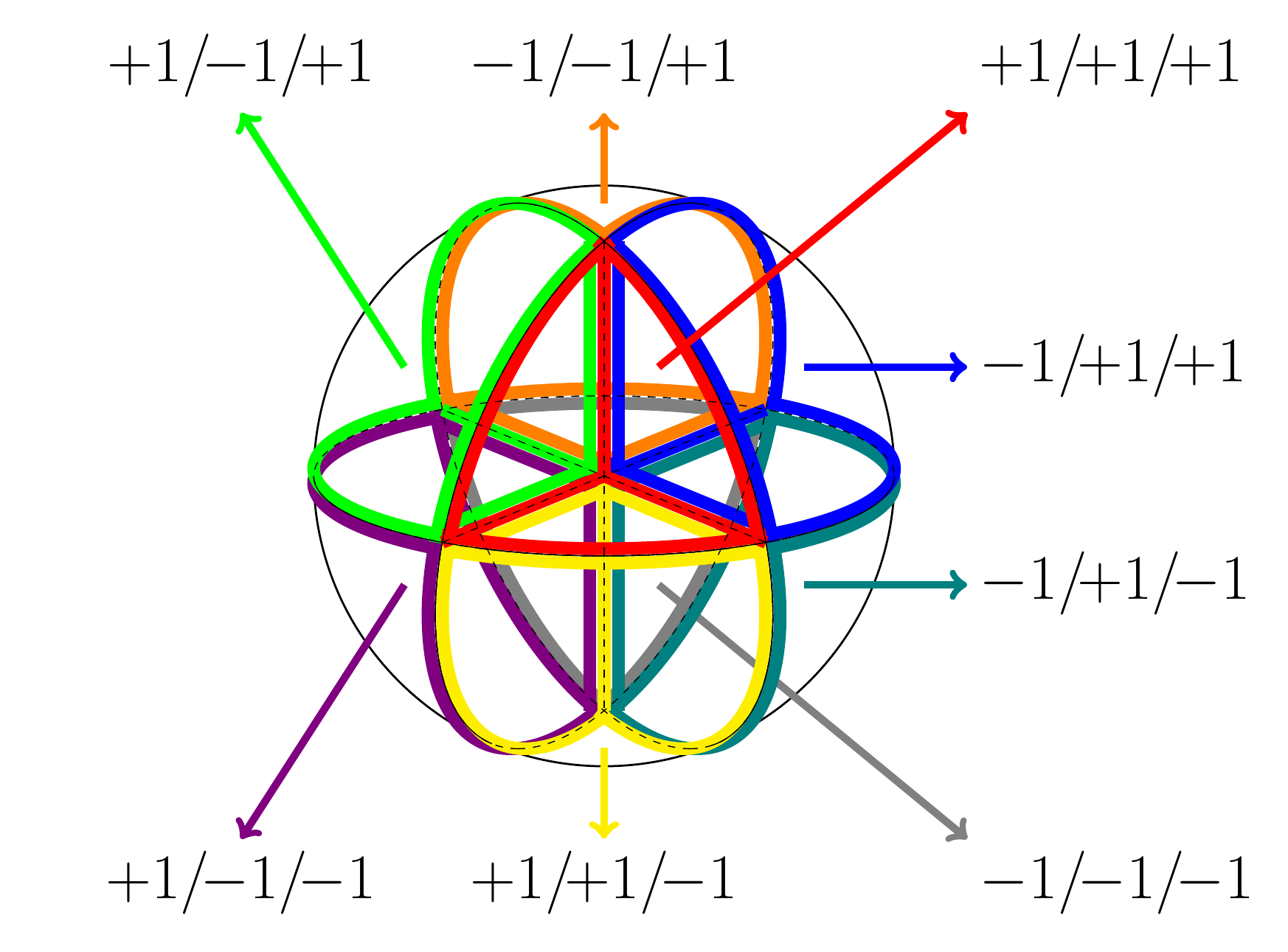}
\includegraphics[trim=15 0 15 0, scale=0.27]{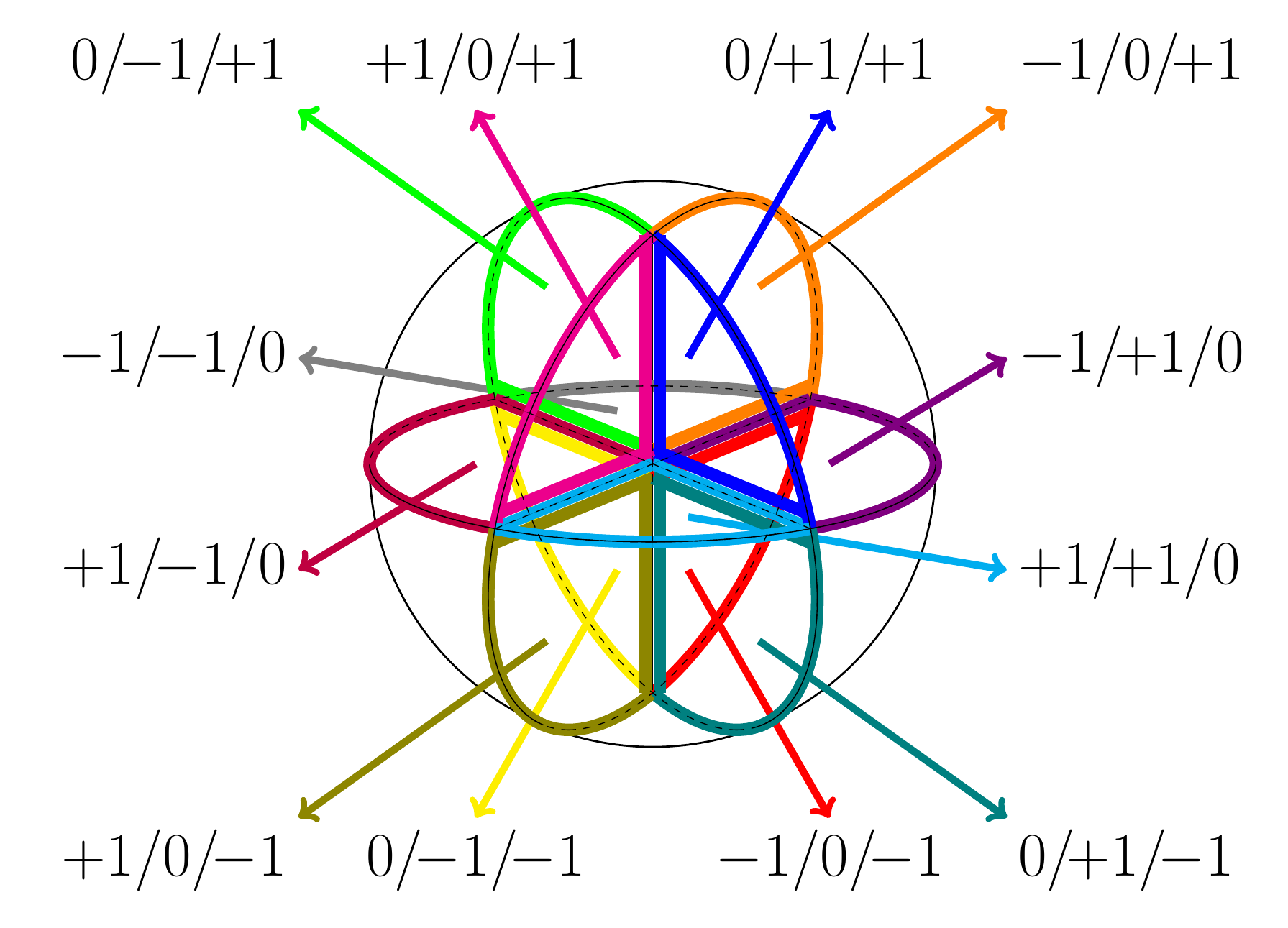}
\includegraphics[scale=0.27, trim=0 0 130 0]{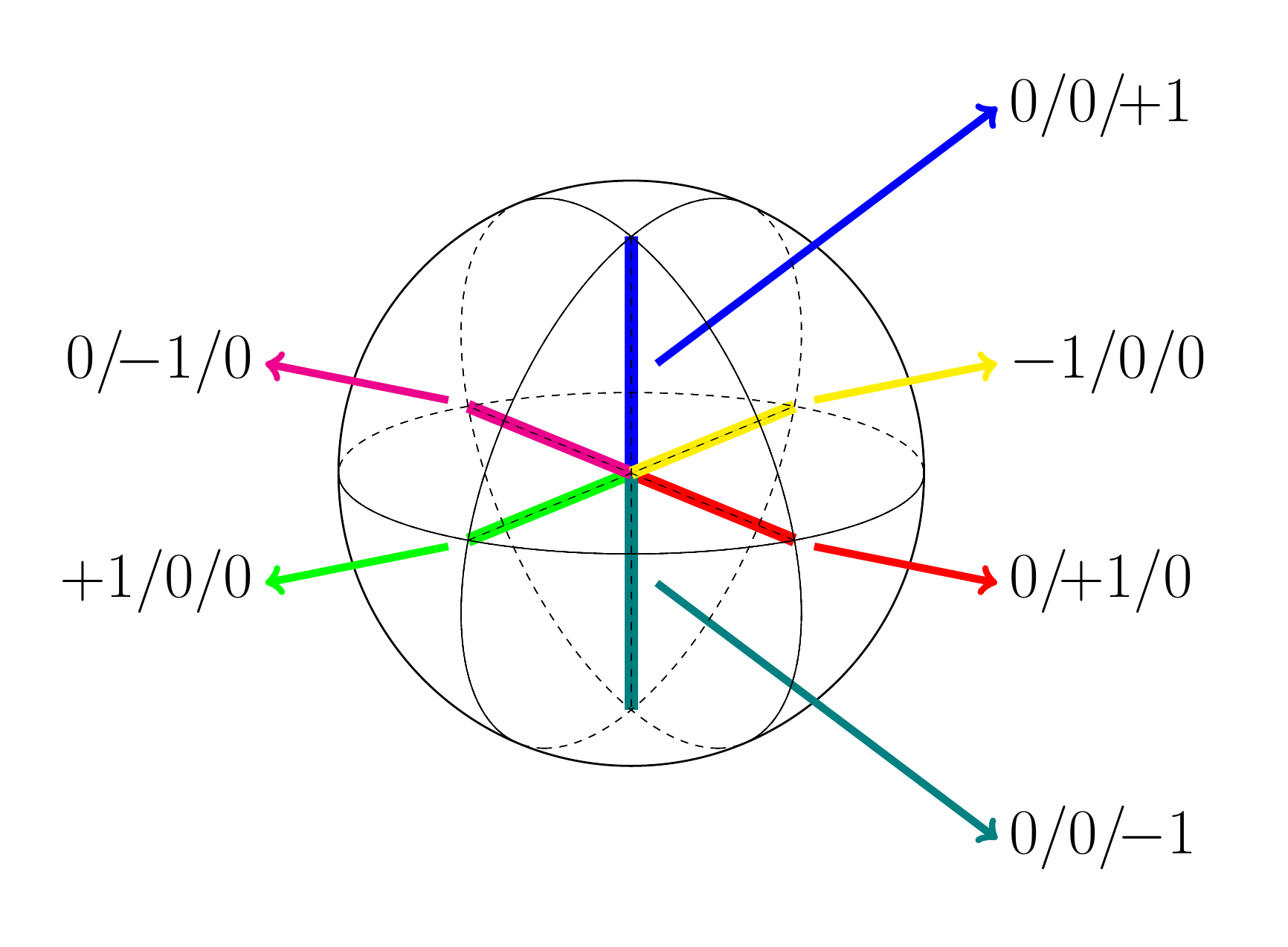}
\end{center}
\caption{Partial ordering of normal vectors. The numbers indicate the signs of normal vector components, which are used in the partial order function. The figures from left to right visualize different (ordered) sign combinations. Sets of sign combinations without zero values form volumes (left), sets with one zero value form planar areas (middle), and sets with two zero values form line sections (right). The 0/0/0 combination forms a point at the origin. }
\label{signBalls}
\end{figure}

\textbf{Learning particle - boundary interactions. } In granular flow simulations, usually particle - particle interaction data outweigh particle - boundary interaction data, which makes learning particle - boundary interactions difficult.
We therefore 
put more emphasis on the type and normal vector features by introducing hyperparameters for the features, that allow discrimination between real particles and virtual boundary particles.
The GNN predictions should be independent of the orientation of the normal vectors representing the planes of the triangle walls. Since there is a positive and a negative choice, we use both a positively and a negatively oriented version of the respective normal vectors as input features. However, as the prediction of a network should be invariant with respect to the orientation of the normal vectors, we define a partial ordering to be able to sort the normal vectors with respect to their orientations. For a given normal vector $\Bn=\PAR{n_1,n_2,n_3} \in \R^3$, we use the following partial order function
\begin{equation}
f_o\PAR{\Bn}=\sum_{i=1}^{3} 3^{i-1}\;\PAR{\sgn\PAR{n_i}+1}
\end{equation}
to retrieve the scalar values $o_1=f_o\PAR{\Bn}$ and $o_2 = f_o\PAR{-\Bn}$ and sort the two vectors according to their corresponding mapped values. 
We choose a base of $3$ since zero entries of vectors are  possible. Different sign combinations of the normal vectors are shown in Figure~\ref{signBalls}.

To test the performance of our approach, we conduct a toy experiment as well as a simulation experiment with different representations of normal vectors. We describe both experiments in the appendix (Section \ref{toyExp} and Section \ref{simExp}).

\section{Results}

We focus on two typical applications of granular flow simulations in the design of industrial machinery: the particle flow through a hopper and the particle dynamics in a rotating drum.
Figure \ref{paravisplot} visualizes
the particle positions for both problem settings at different
time steps in the ground truth simulation and in the prediction of our model.
For both applications gravitation acts along the z-direction.
The upper part of the hopper is delimited along the y-axis by two planes, which are parallel to the x-z plane. The x-axis is delimited by two planes, that are inclined at certain angles $\alpha,\, 180^{\circ}-\alpha$ to the x-y plane and at corresponding angles $\alpha-90^{\circ}, 90^{\circ}-\alpha$ to the y-z plane. At the bottom of the hopper there is a hole with an adjustable radius, that is initially closed. Our generated training data consists of 30 simulation trajectories with different angles $\alpha$ and different hole sizes. Moreover, the initial filling distribution is varied.
The rotation axis of the drum is the y-axis. The initial filling is obtained by rotating the filling of a resting drum around the x-axis.

\begin{figure}[H]
	\centering
	\makebox[ \textwidth ]{
	\includegraphics[trim=0 100 0 0,clip, height=130px]{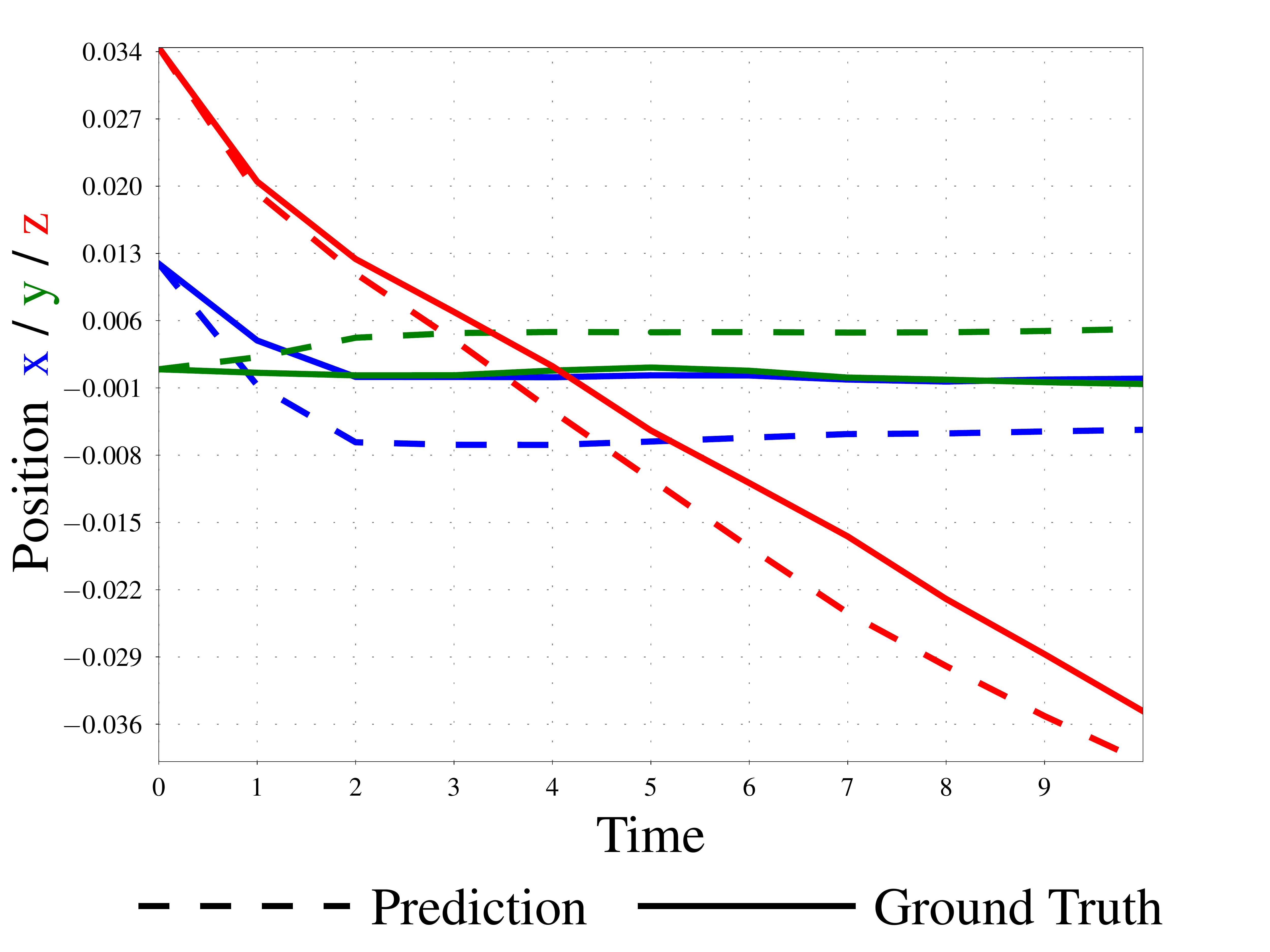}
	\includegraphics[trim=0 100 0 0,clip,height=130px]{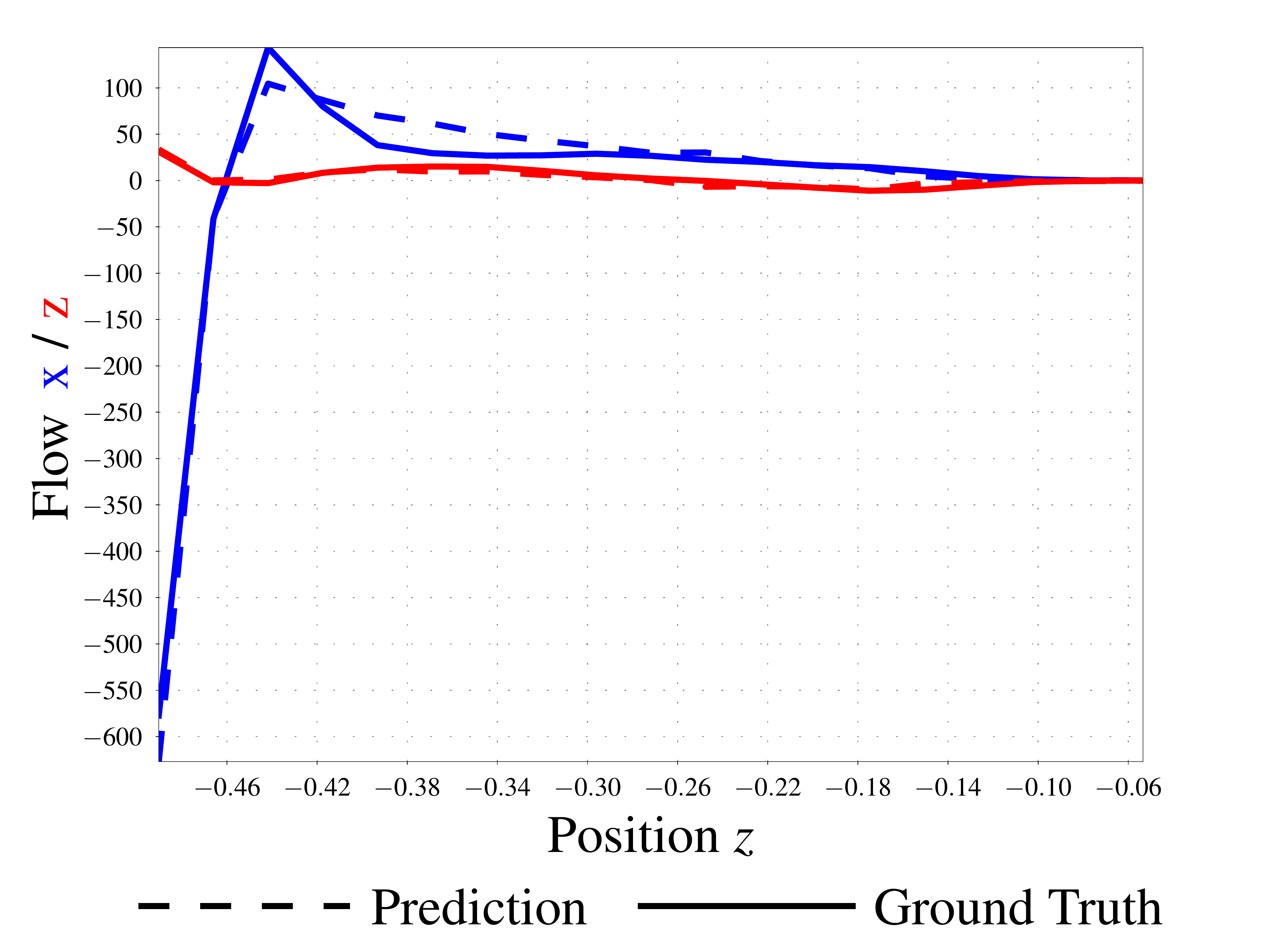}
	}
	\makebox[ \textwidth ]{
    \includegraphics[trim=0 100 0 0,clip,height=130px]{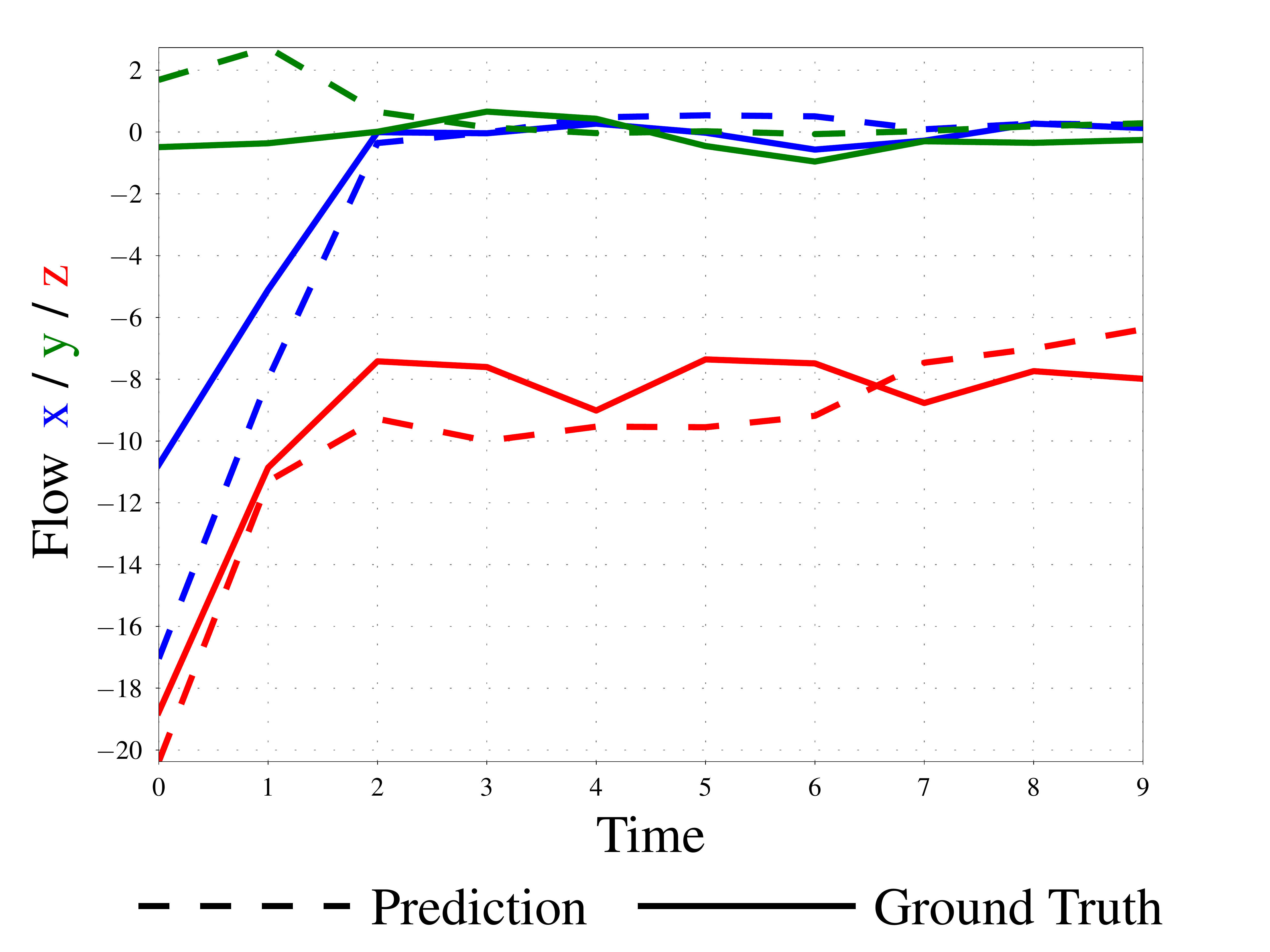}
	\includegraphics[trim=0 100 0 0,clip,height=130px]{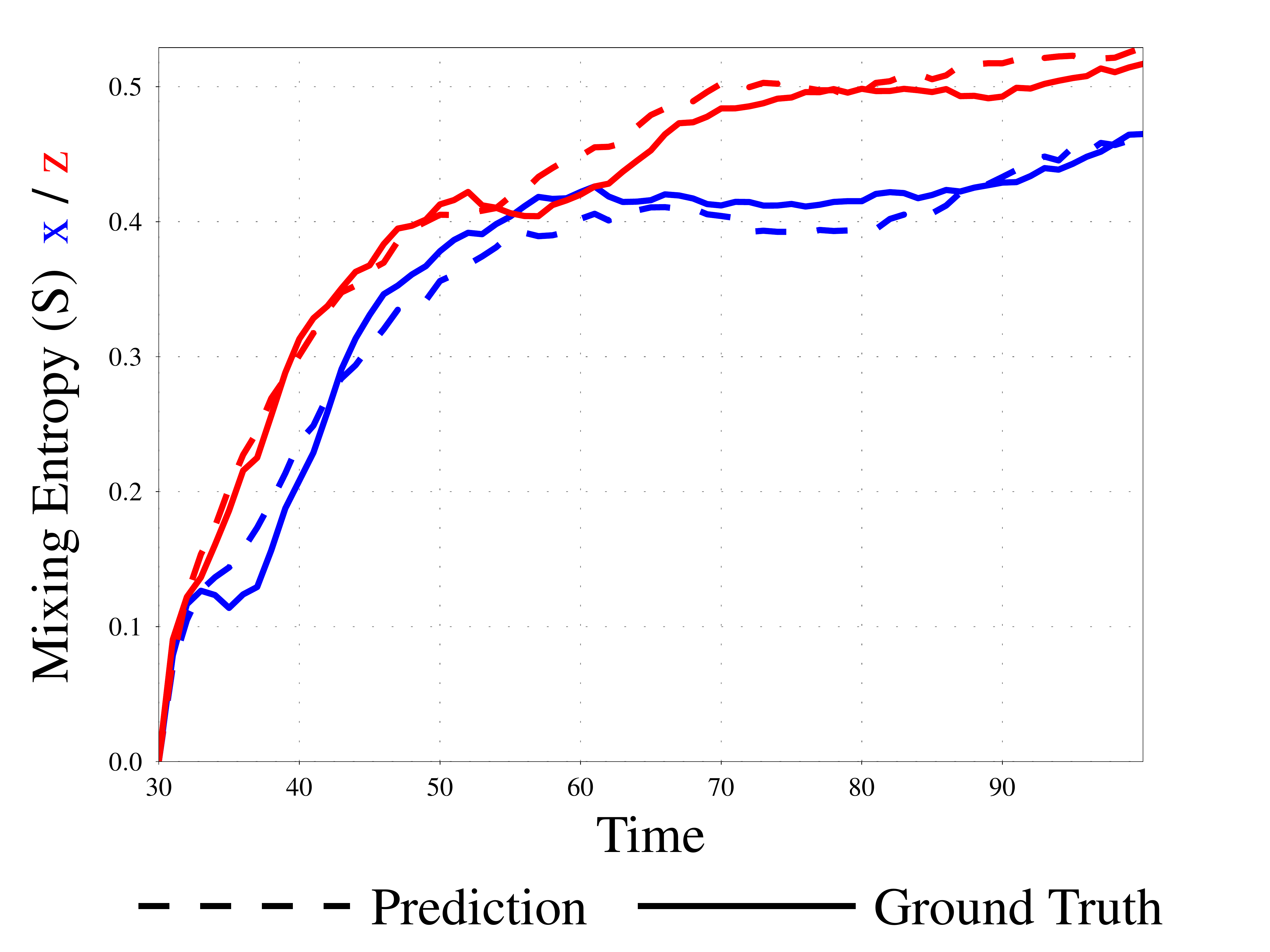}
	}
	\makebox[ \textwidth ]{
    \includegraphics[trim=0 0 0 990,clip,height=12px]{figuresSIMDL/flow_hop.pdf}
    }
	\caption{Position (upper left) and flow profile (lower left) plots for the Hopper, and, flow profile (upper right) and entropy plot (lower right) for the Drum. The plots visualize ground truth (solid line) vs. predictions (dashed line) in dependence of the time step or a coordinate. The flow profiles are average velocities of of particles at a given time step or $z$ coordinate. The mixing entropies are obtained by splitting particles into two partitions according to a threshold on the respective $x$ or $z$ coordinate at time step 30.}	
	\label{cmpplot}
\end{figure}

In Figure \ref{cmpplot} we compare ground truth trajectories to the learned trajectories.
The machine learning model does not exactly reproduce the ground truth trajectories due to chaotic behavior in the long run, but aggregated quantities like particle-averaged positions (upper left plot) show good qualitative agreement. The upper right and the lower left plots of the figure show that also the time- and particle-averaged particle flows are well reproduced. In the lower right plot we analyse the machine learning model in terms of mixing behavior. We quantify the extend of particle mixing via the mixing entropy \citep{fang1975appl}, which is computed by splitting particles into two classes +1, -1 at a certain time step $t_0=30$ and then summing over weighted local entropies $s(\Bx_{klm}, t)$ located at grid cells $\Bx_{klm}$ (see Appendix \ref{mixent} for details).  The lower right plot shows that the expected increase of the mixing entropy over time is well described by the GNN.
Overall, we conclude that our GNN approach is able to model 3D granular flow simulations accurately.

\section{Outlook}

Although, we have already used comparably large time steps in our machine learning model (i.e. multiples
of the ground truth simulation training data), we have not fully investigated the potential
of machine learning models for granular simulation data yet: Firstly, the simulation time of the ground truth simulation is strongly material-dependent. As a downside, the true Young‘s Modulus, which describes a stress-strain relationship in the linear elastic region of a material, is usually not used in the computation of simulations, but a much smaller value.
Machine learning models therefore could allow the simulation of granular flows with more realistic material parameters.
Secondly, larger time steps would allow to model physical phenomena occurring on larger time scales, which cannot be modeled accurately with current DEM approaches.
Further, we plan to leverage the
multiple symmetries occurring in the geometry of granular flow problems.
Another research direction might be the usage of global parameters
such as the angles of the side walls of the hopper as additional input features to the neural network.

\subsubsection*{Acknowledgments}
This research was supported by FFG grant 871302 (DL for granular flow). 

The ELLIS Unit Linz, the LIT AI Lab, the Institute for Machine Learning, are supported by the Federal State Upper Austria. IARAI is supported by Here Technologies. We thank the projects AI-MOTION (LIT-2018-6-YOU-212), DeepToxGen (LIT-2017-3-YOU-003), AI-SNN (LIT-2018-6-YOU-214), DeepFlood (LIT-2019-8-YOU-213), Medical Cognitive Computing Center (MC3), PRIMAL (FFG-873979), S3AI (FFG-872172), ELISE (H2020-ICT-2019-3 ID: 951847), AIDD (MSCA-ITN-2020 ID: 956832).
We thank Janssen Pharmaceutica (MaDeSMart, HBC.2018.2287), Audi.JKU Deep Learning Center, TGW LOGISTICS GROUP GMBH, Silicon Austria Labs (SAL), FILL Gesellschaft mbH, Anyline GmbH, Google, ZF Friedrichshafen AG, Robert Bosch GmbH, UCB Biopharma SRL, Merck Healthcare KGaA, Software Competence Center Hagenberg GmbH, T\"{U}V Austria, and the NVIDIA Corporation.

\clearpage

\vspace*{-2cm}

\imagelabelset{
coarse grid color = red,
fine grid color = gray,
image label font = \sffamily\small,
image label distance = 2mm,
image label back = black,
image label text = white,
coordinate label font = \normalfont,
coordinate label distance = 2mm,
coordinate label back = white,
coordinate label text = black,
annotation font = \normalfont\small,
arrow distance = 1.5mm,
border thickness = 0.6pt,
arrow thickness = 0.4pt,
tip size = 1.2mm,
outer dist = 0.5cm,
}

\newsavebox\imagebox

\begin{figure}[h]
	\centering
	\makebox[ \textwidth ]{
	\centering
	\sbox{\imagebox}{
	\begin{annotationimage}[]{trim = 0mm 0mm 0mm 10mm, clip=True, width=0.9\textwidth}{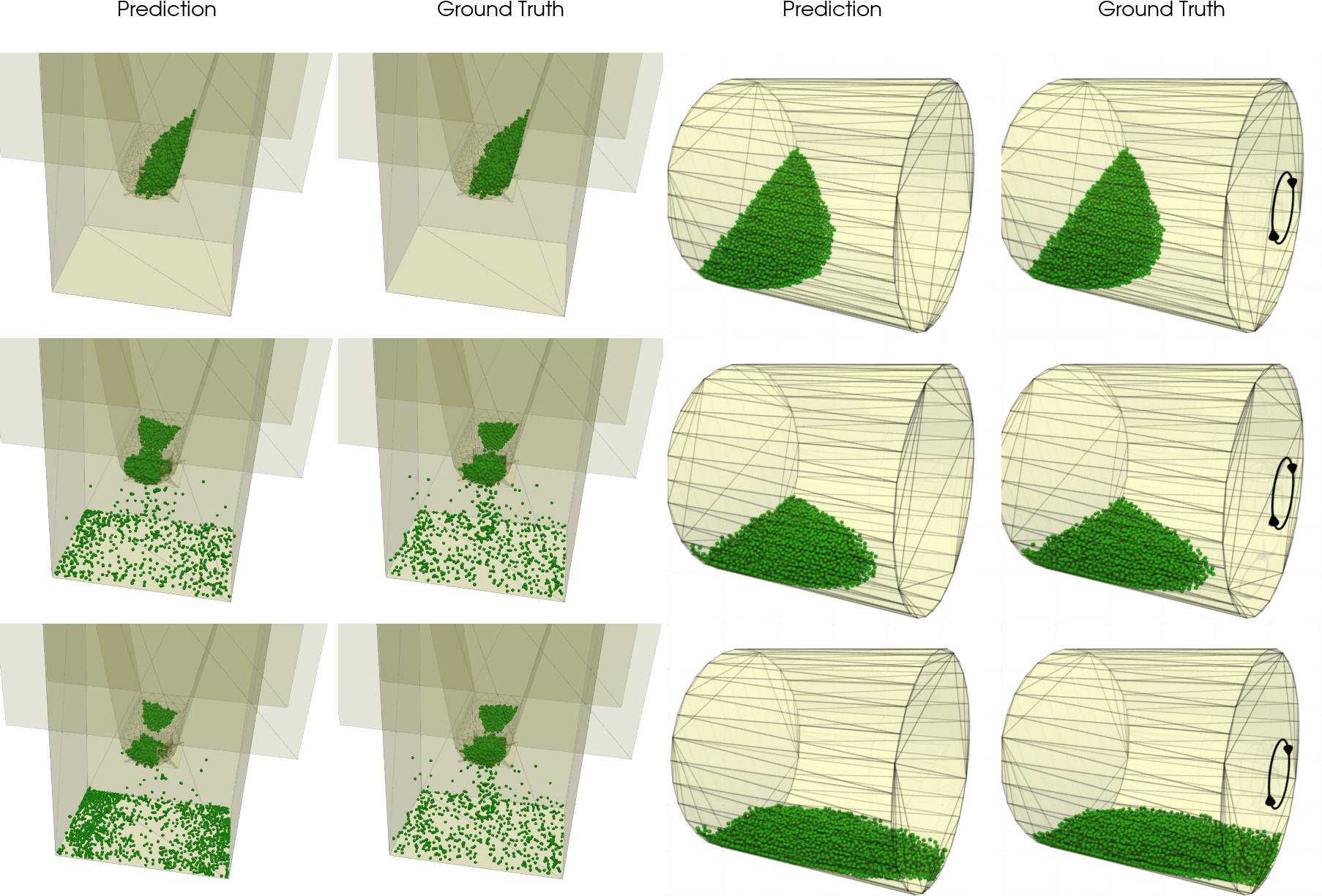}

\draw[coordinate label = {Hopper at (0.25,1.08)}];
\draw[coordinate label = {Rotating Drum at (0.75,1.08)}];
\draw[coordinate label = {Prediction at (0.125,1.03)}];
\draw[coordinate label = {Ground Truth at (0.375,1.03)}];
\draw[coordinate label = {Prediction at (0.625,1.03)}];
\draw[coordinate label = {Ground Truth at (0.875,1.03)}];
    \draw [thick,->] (-0.04, 1)  -- node[anchor=south, rotate=90] {Time} (-0.04, 0);
	\end{annotationimage}
	}
	\usebox{\imagebox}
	}

	\caption{Particle distributions for Hopper and Drum dynamics. Data obtained by the particle simulator LIGGGHTS (Ground Truth) and our trained graph neural network (Prediction) are compared. Particles are indicated by green spheres, triangular wall areas are yellow, the edges of these triangles are indicated by grey lines. The circular arrow indicates the rotation direction of the Drum. }	
	\label{paravisplot}
\end{figure} 

\clearpage

\bibliography{iclr2021_conference}
\bibliographystyle{iclr2021_conference}

\appendix

\numberwithin{equation}{section}
\numberwithin{figure}{section}
\numberwithin{table}{section}

\section{Discrete element Method (DEM)}
\label{DEM}

The key idea of of the Discrete element Method (DEM) is that the granular medium is represented by discrete objects, commonly referred to as particles (e.g. spheres or polyhedra) and that they interact by exchanging momentum at a particle - particle contact level using a so-called contact model.
The most basic contact model is a spring-dashpot model, in which the interaction force $F_{ij}$ between two particles $i$ and $j$ is given as
\begin{equation}
    F_{ij} = k \delta_{ij} - \gamma v_{ij} \ ,
\end{equation}
where $k$ is the spring stiffness, $\delta_{ij}$ is the overlap of the two particles, $\gamma$ is the damping constant and $v_{ij}$ is the relative velocity between the two particles. Such contact models can become prohibitively complex in order to model phenomena such as cohesion \citep{obermayr2014}, surface roughness and others.
One significant downside of the DEM approach is that the time integration of the Lagrangian particles requires small time steps to properly resolve the particle contacts. In industrial applications there often is the need to study physical phenomena which occur on different time scales, e.g. particle collisions ($\mathcal{O}(10^{-5}s)$) vs. moisture content in particles ($\mathcal{O}(1s)$, \citet{mellmann2011}), which can lead to weeks of simulation time. While advances have been made to overcome such issues (e.g. \citet{Kloss2017}), they remain limited in their application, due to the fact that they rely on prior simulation of the exact setup and cannot be used for interpolation of quantities directly related to the flow behavior.

\section{Normal Vector Representations}

\subsection{Reflection Toy Example}
\label{toyExp}

We conduct a toy experiment to showcase that a canonical representation of normal vectors is helpful for learning 3D simulations. In detail, 
we consider reflection at a plane as given by $Ref_{\Bn}\left(\Bv\right) = \Bv - 2 \frac{\Bv \cdot \Bn}{\Bn \cdot \Bn} \Bn$ and try to learn the reflection formula by a simple ReLU network, which takes the 3 components of $\Bn$ and $\Bv$ as input features and predicts the 3 components of $Ref_{\Bn}\left(\Bv\right)$. The training data thereby consists of reflections at four fixed walls: the top, the bottom, the left, and, the right side of a simple cube. 

\begin{figure}[H]
	\centering
	\makebox[ \textwidth ]{
	\begin{annotationimage}[]{trim = 0mm 0mm 0mm 0mm, clip=False, height=140px}{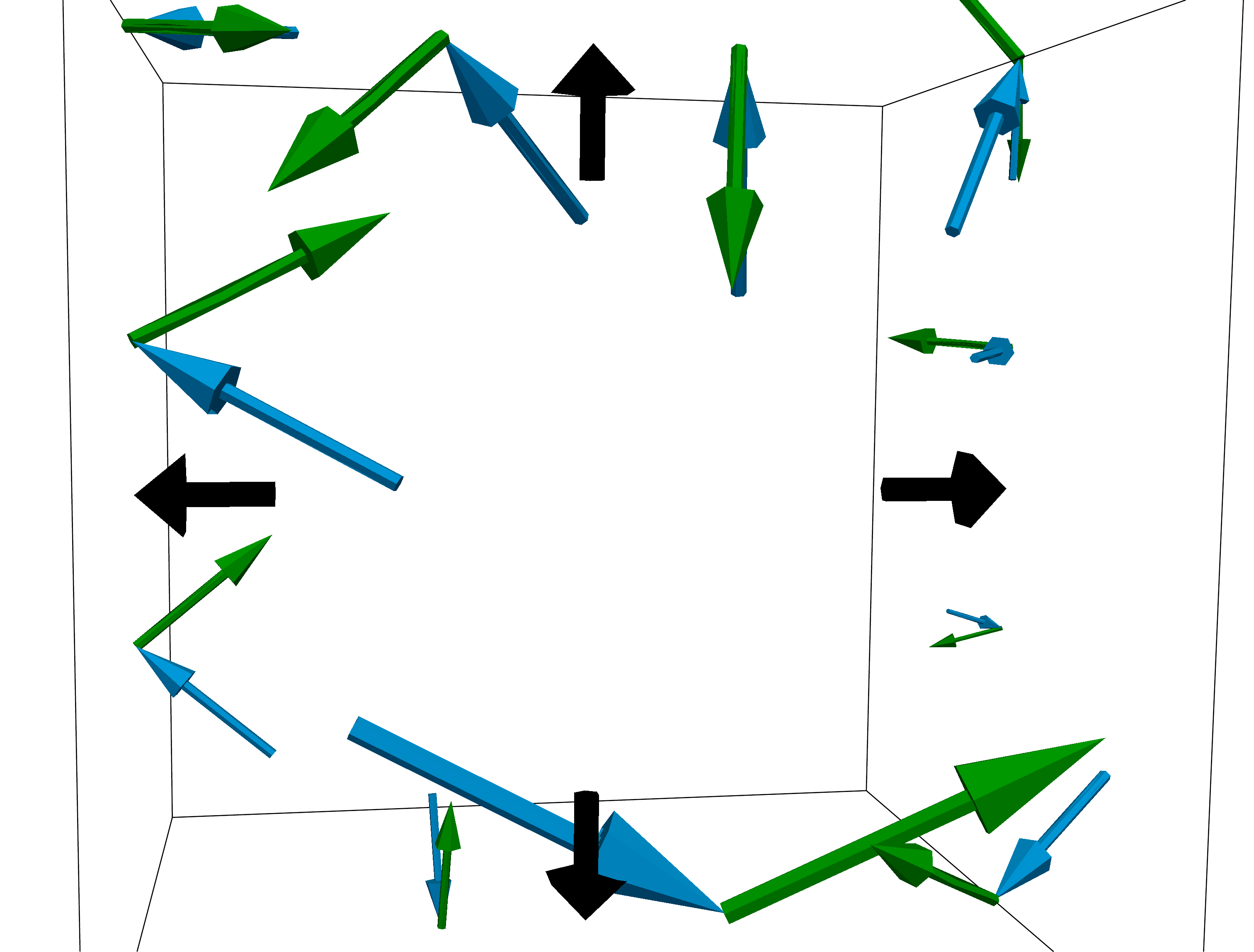} 
	\draw[coordinate label = {Ground Truth: $\Bn$  at (0.5,1.08)}];
	\end{annotationimage}
	\begin{overpic}[scale=0.1, tics=5, trim=-1cm -20cm 0 0cm, clip]{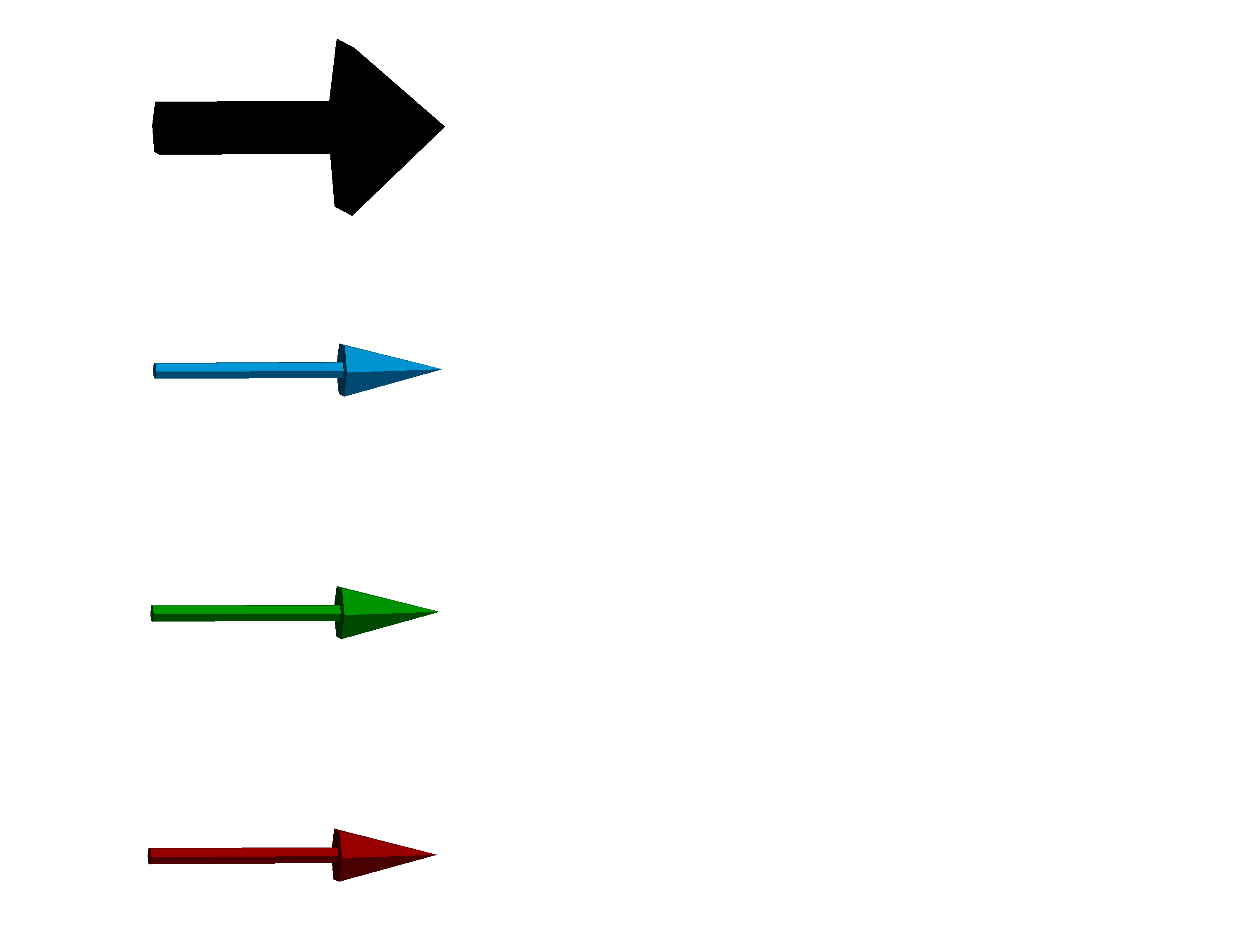}
	\put(30,91){Wall Normal Vectors}
	\put(30,77){Incident Rays}
	\put(30,63){True Reflected Rays}
	\put(30,49){Predicted Reflected Rays}
	\end{overpic}
	}\\
    \vspace{0.5cm}
	\makebox[ \textwidth ]{
	\begin{annotationimage}[]{trim = 0mm 0mm 0mm 0mm, clip=False, height=140px}{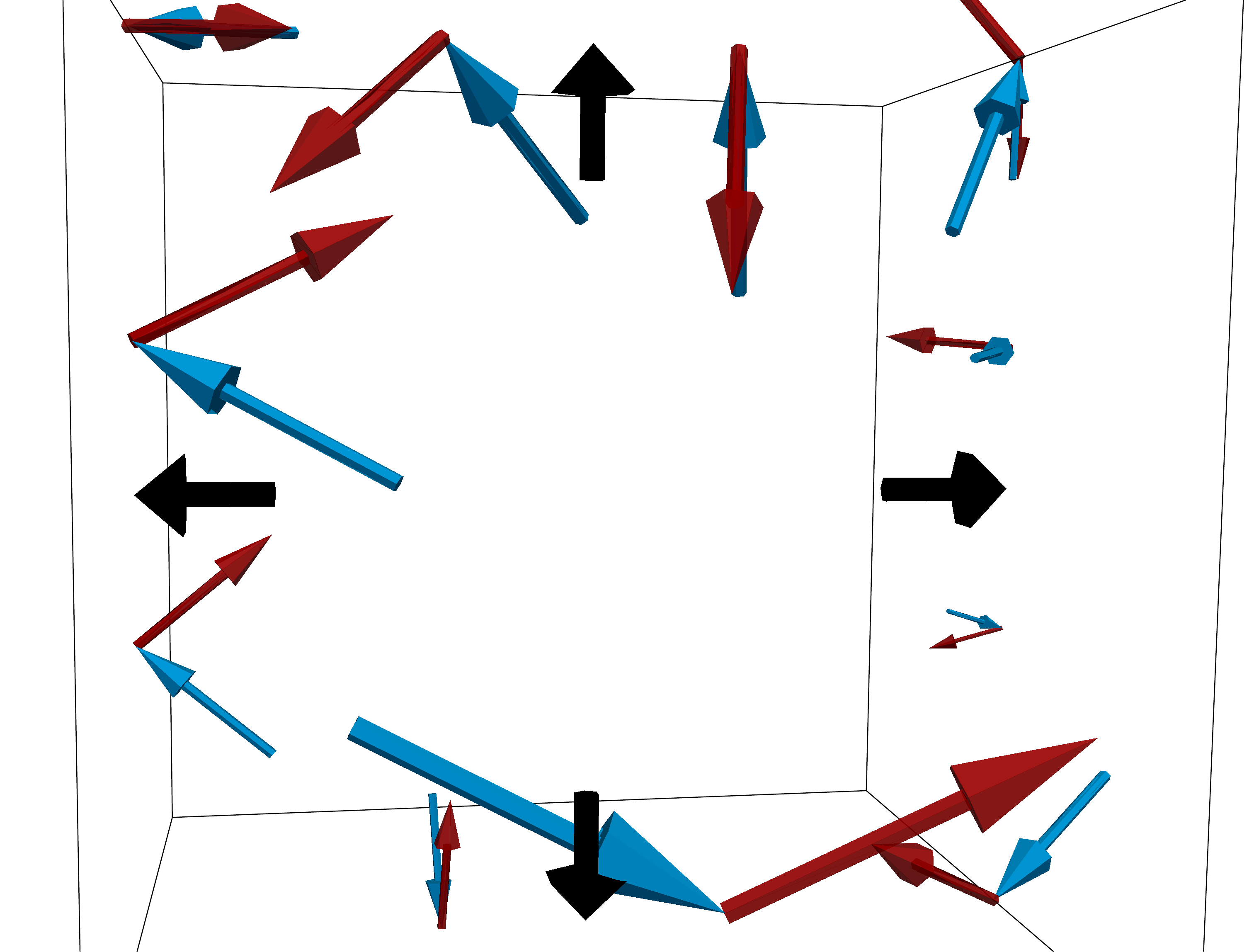}
	\draw[coordinate label = {R1: $\Bn$ at (0.5,1.08)}];
	\end{annotationimage}
	\begin{annotationimage}[]{trim = 0mm 0mm 0mm 0mm, clip=False, height=140px}{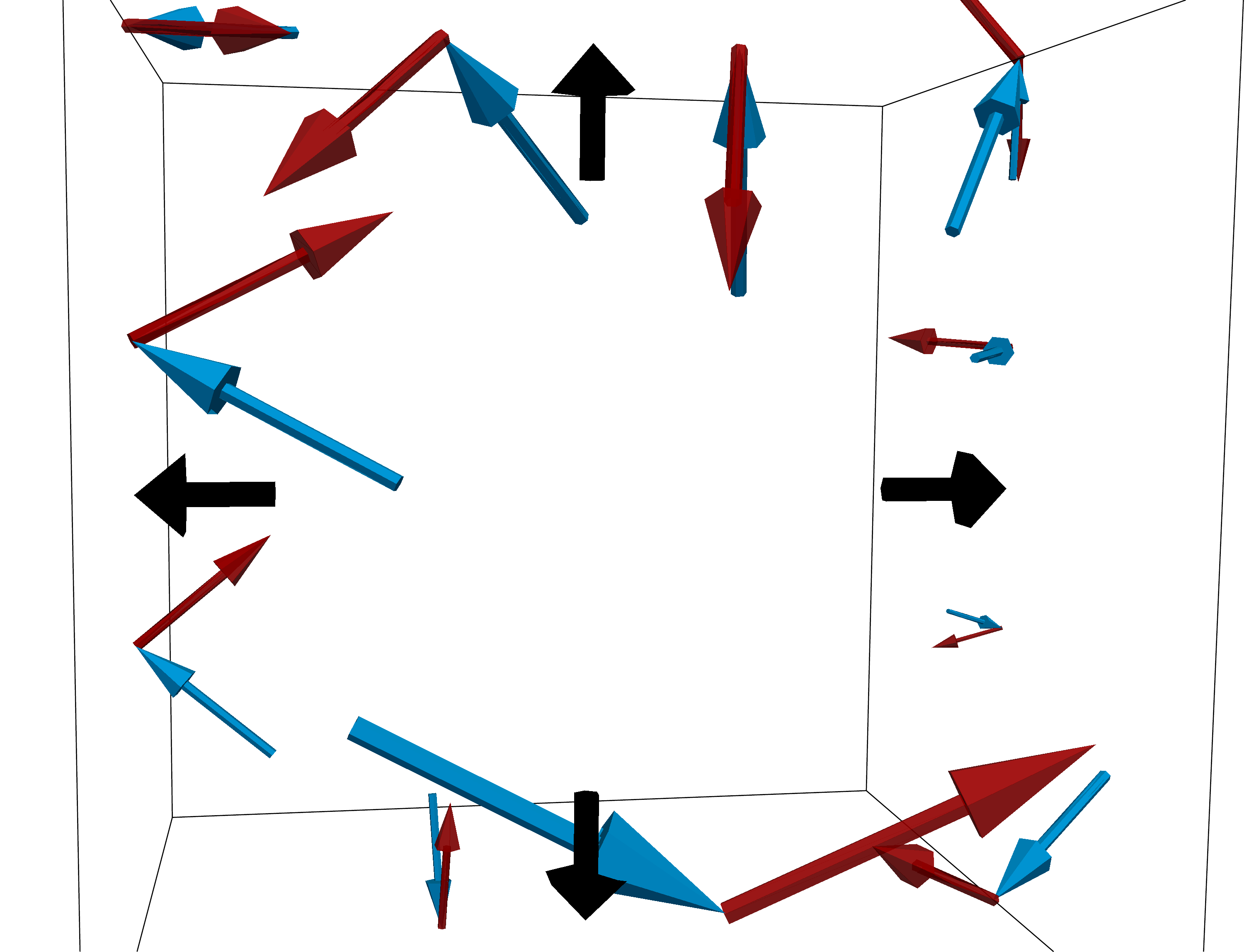} 
	\draw[coordinate label = {R2: $\Bn, -\Bn$ at (0.5,1.08)}];
	\end{annotationimage}
	}
	\\
	\vspace{0.5cm}
	\makebox[ \textwidth ]{
	\begin{annotationimage}[]{trim = 0mm 0mm 0mm 0mm, clip=False, height=140px}{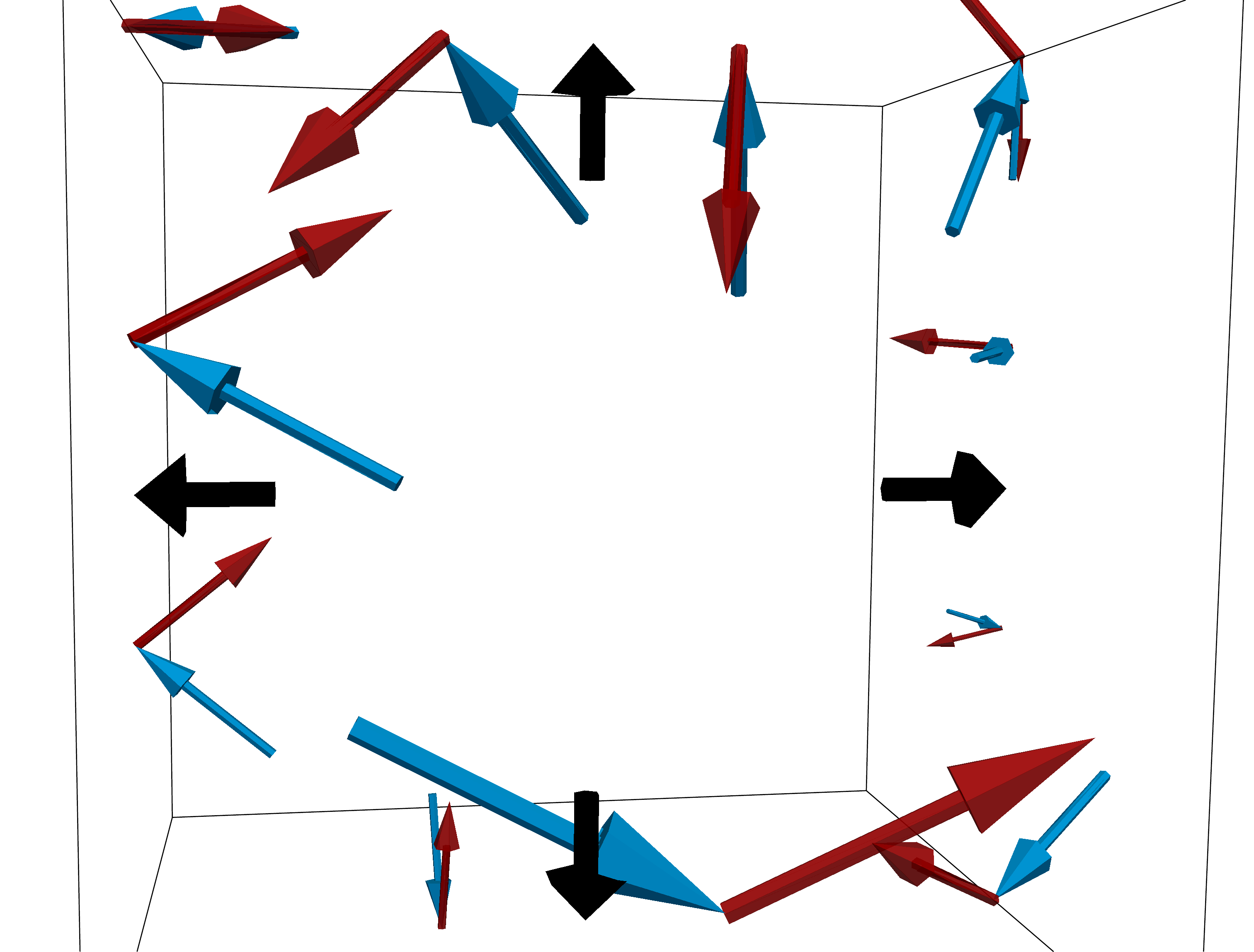} 
	\draw[coordinate label = {R3: $\left\{
	\begin{array}{lc}
        \Bn & \text{if } \mathop{\mathbb{\mathrm{f_o}}}\PAR{\Bn} \leq \mathop{\mathbb{\mathrm{f_o}}}\PAR{-\Bn}\\
        -\Bn & \text{otherwise} 
    \end{array}
    \right.$ at (0.5,1.08)}];
	\end{annotationimage}
	\begin{annotationimage}[]{trim = 0mm 0mm 0mm 0mm, clip=False, height=140px}{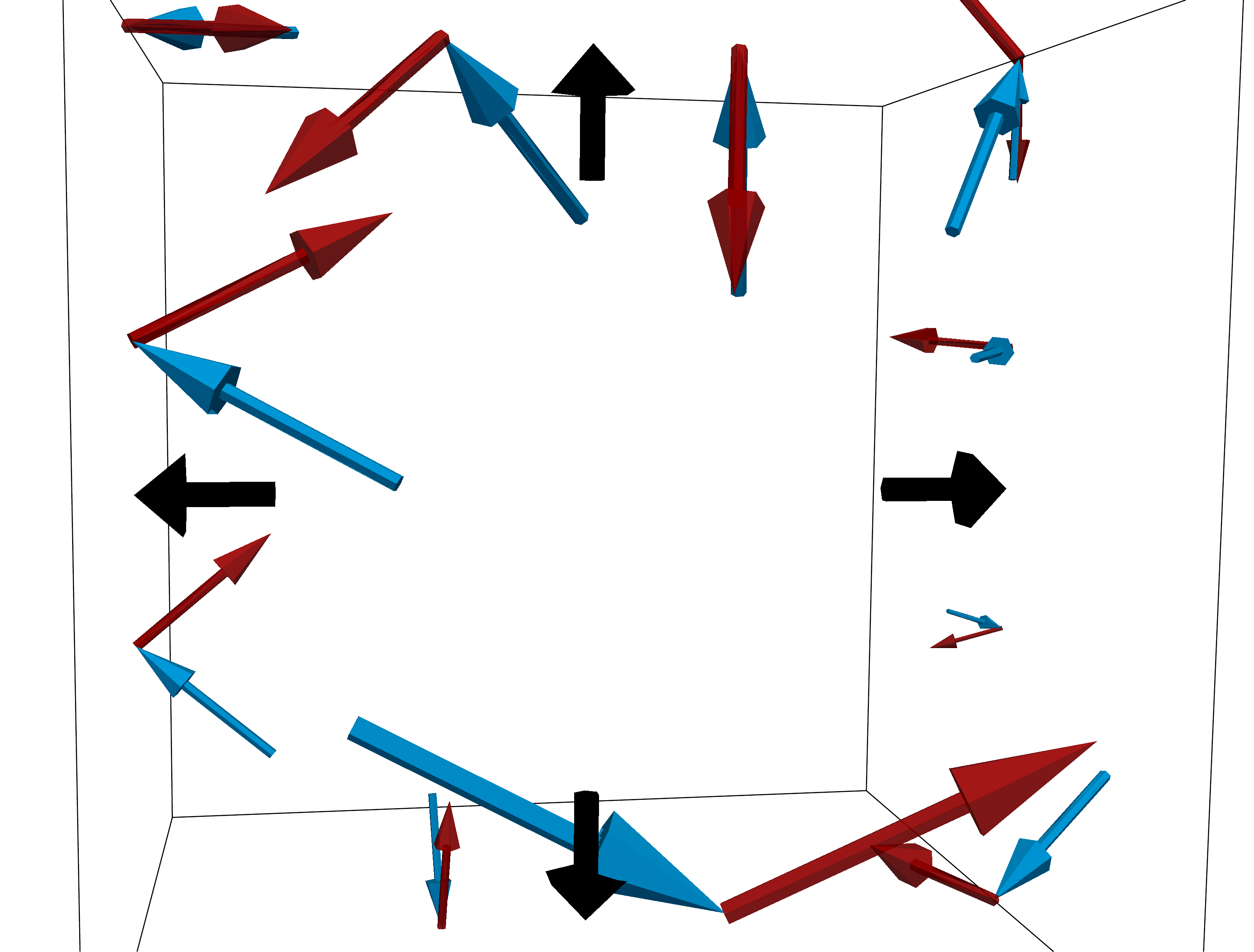} 
	\draw[coordinate label = {R4: $\left\{
	\begin{array}{lc}
        \Bn,-\Bn & \text{if } \mathop{\mathbb{\mathrm{f_o}}}\PAR{\Bn} \leq \mathop{\mathbb{\mathrm{f_o}}}\PAR{-\Bn}\\
        -\Bn,\Bn & \text{otherwise} 
    \end{array}
    \right.$ at (0.5,1.08)}];
    \end{annotationimage}
	}
	\caption{\label{reflTrain} Reflection of rays at four different walls (left, right, bottom, top). Wall normal vectors are visualized by black arrows. The incident rays are visualized by blue arrows, reflected rays are indicated by green arrows in the ground truth plot. Red arrows in plots R1-R4 visualize neural
	network predictions. Neural network predictions are based on wall representations \textbf{that are oriented the same way as in the training phase}. The caption above each plot indicates the wall input features used by each of the trained networks.}
\end{figure}

\begin{figure}[H]
	\centering
	\makebox[ \textwidth ]{
	\begin{annotationimage}[]{trim = 0mm 0mm 0mm 0mm, clip=False, height=140px}{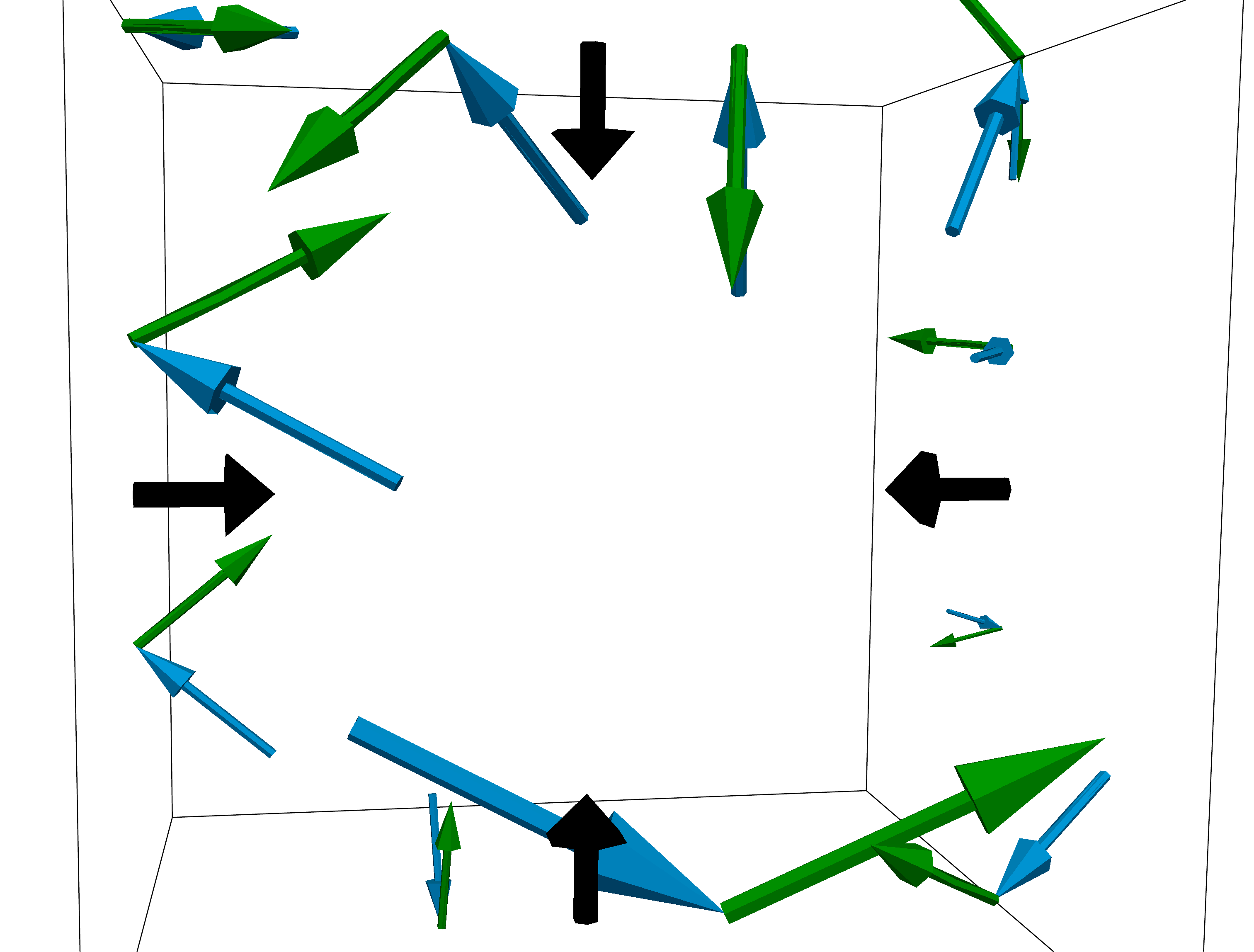} 
	\draw[coordinate label = {Ground Truth: $-\Bn$  at (0.5,1.08)}];
	\end{annotationimage}
	\begin{overpic}[scale=0.1, tics=5, trim=-1cm -20cm 0 0cm, clip]{figuresAdd/demo/arrows.pdf}
	\put(30,91){Wall Normal Vectors}
	\put(30,77){Incident Rays}
	\put(30,63){True Reflected Rays}
	\put(30,49){Predicted Reflected Rays}
	\end{overpic}
	}\\
    \vspace{0.5cm}
	\makebox[ \textwidth ]{
	\begin{annotationimage}[]{trim = 0mm 0mm 0mm 0mm, clip=False, height=140px}{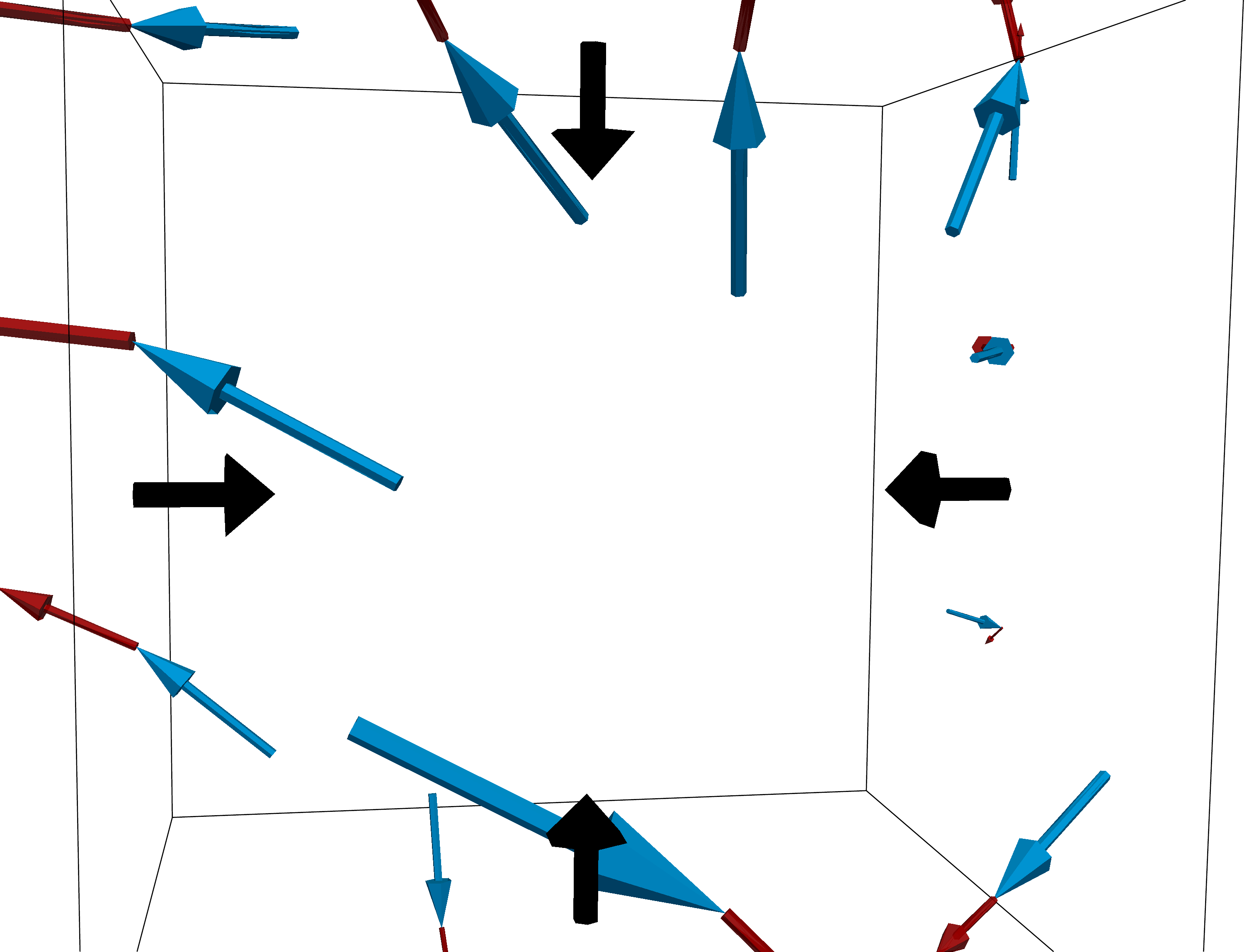}
	\draw[coordinate label = {R1: $-\Bn$ at (0.5,1.08)}];
	\end{annotationimage}
	\begin{annotationimage}[]{trim = 0mm 0mm 0mm 0mm, clip=False, height=140px}{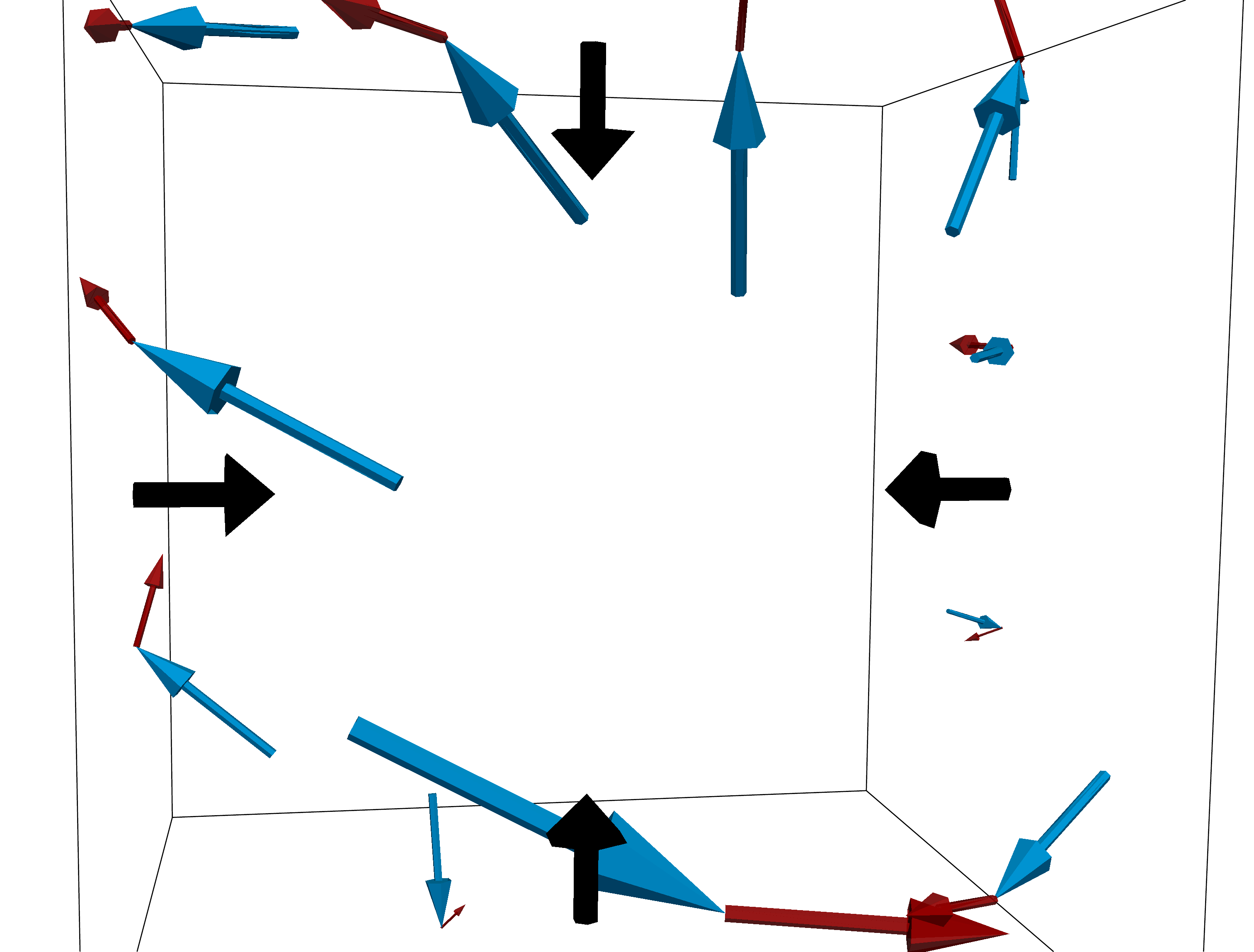} 
	\draw[coordinate label = {R2: $-\Bn, \Bn$ at (0.5,1.08)}];
	\end{annotationimage}
	}\\
	\vspace{0.5cm}
	\makebox[ \textwidth ]{
	\begin{annotationimage}[]{trim = 0mm 0mm 0mm 0mm, clip=False, height=140px}{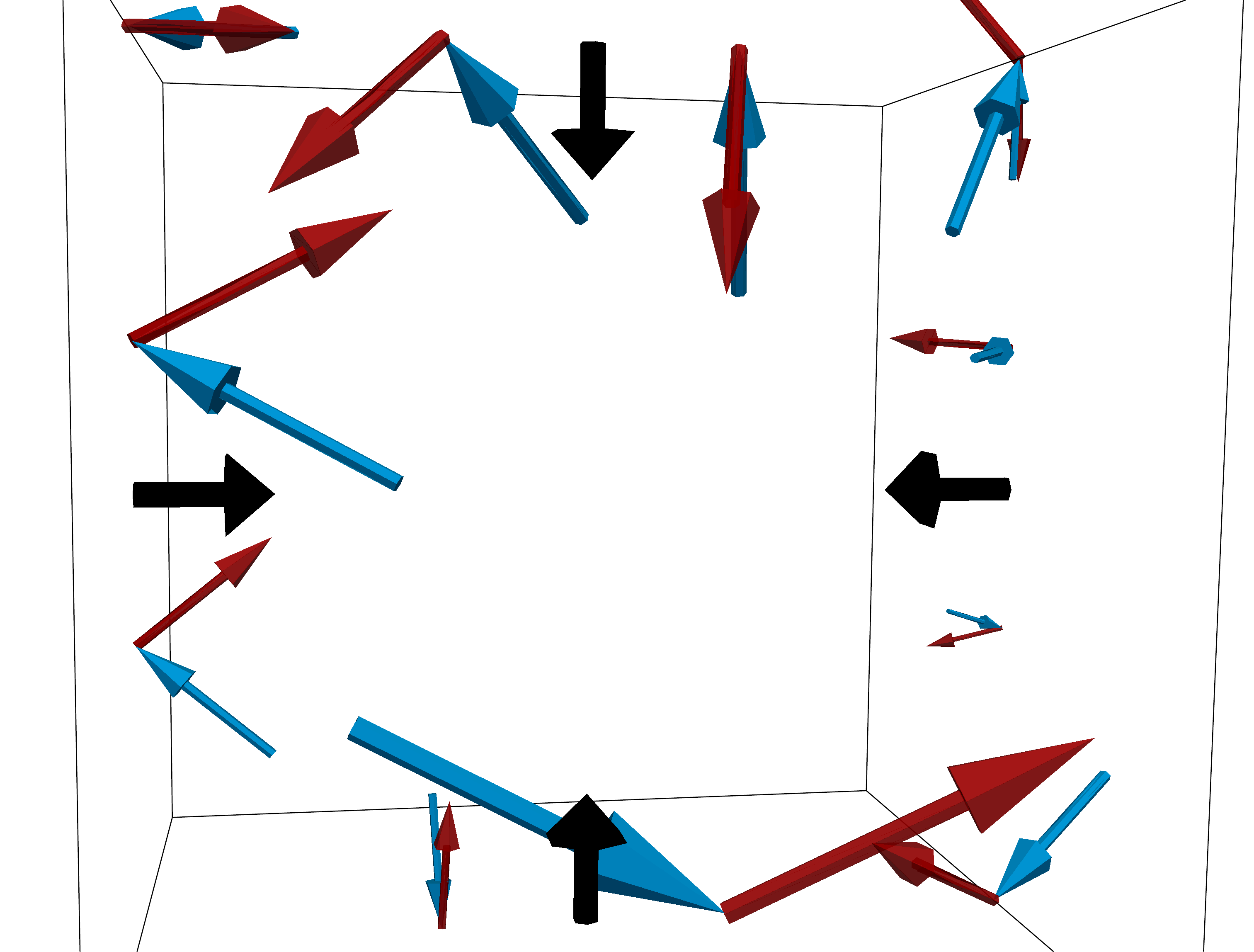} 
	\draw[coordinate label = {R3: $\left\{
	\begin{array}{lc}
        -\Bn & \text{if } \mathop{\mathbb{\mathrm{f_o}}}\PAR{-\Bn} \leq \mathop{\mathbb{\mathrm{f_o}}}\PAR{\Bn}\\
        \Bn & \text{otherwise} 
    \end{array}
    \right.$ at (0.5,1.08)}];
	\end{annotationimage}
	\begin{annotationimage}[]{trim = 0mm 0mm 0mm 0mm, clip=False, height=140px}{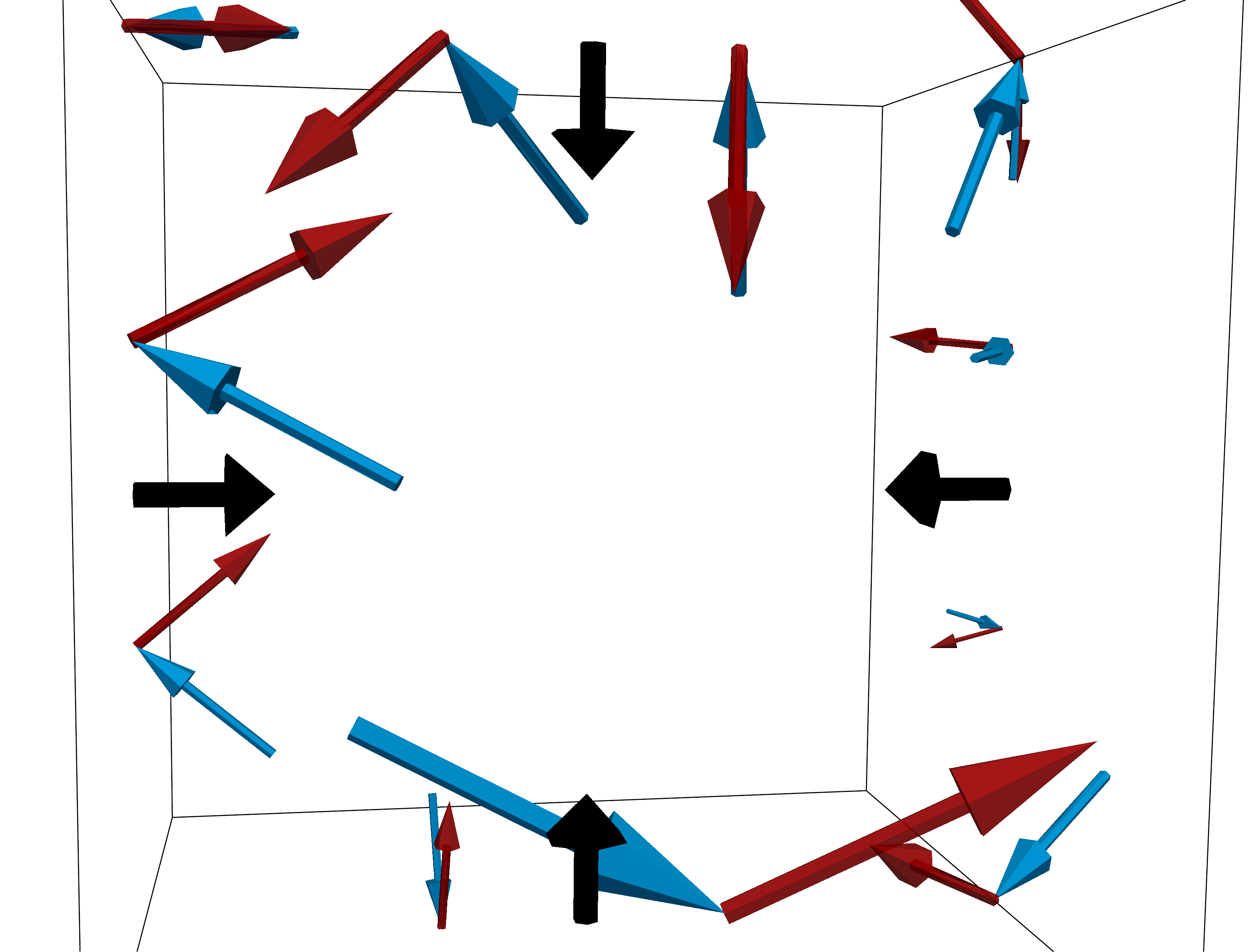} 
	\draw[coordinate label = {R4: $\left\{
	\begin{array}{lc}
        -\Bn,\Bn & \text{if } \mathop{\mathbb{\mathrm{f_o}}}\PAR{-\Bn} \leq \mathop{\mathbb{\mathrm{f_o}}}\PAR{\Bn}\\
        \Bn,-\Bn & \text{otherwise} 
    \end{array}
    \right.$ at (0.5,1.08)}];
    \end{annotationimage}
	}
	\caption{\label{reflTrainInv} Reflection of rays at four different walls (left, right, bottom, top). Wall normal vectors are visualized by black arrows. The incident rays are visualized by blue arrows, reflected rays are indicated by green arrows in the ground truth plot. Red arrows in plots R1-R4 visualize neural
	network predictions. Neural network predictions are based on wall representations \textbf{that are inversely oriented compared to the training phase}. The caption above each plot indicates the wall input features used by each of the trained networks.}
\end{figure}
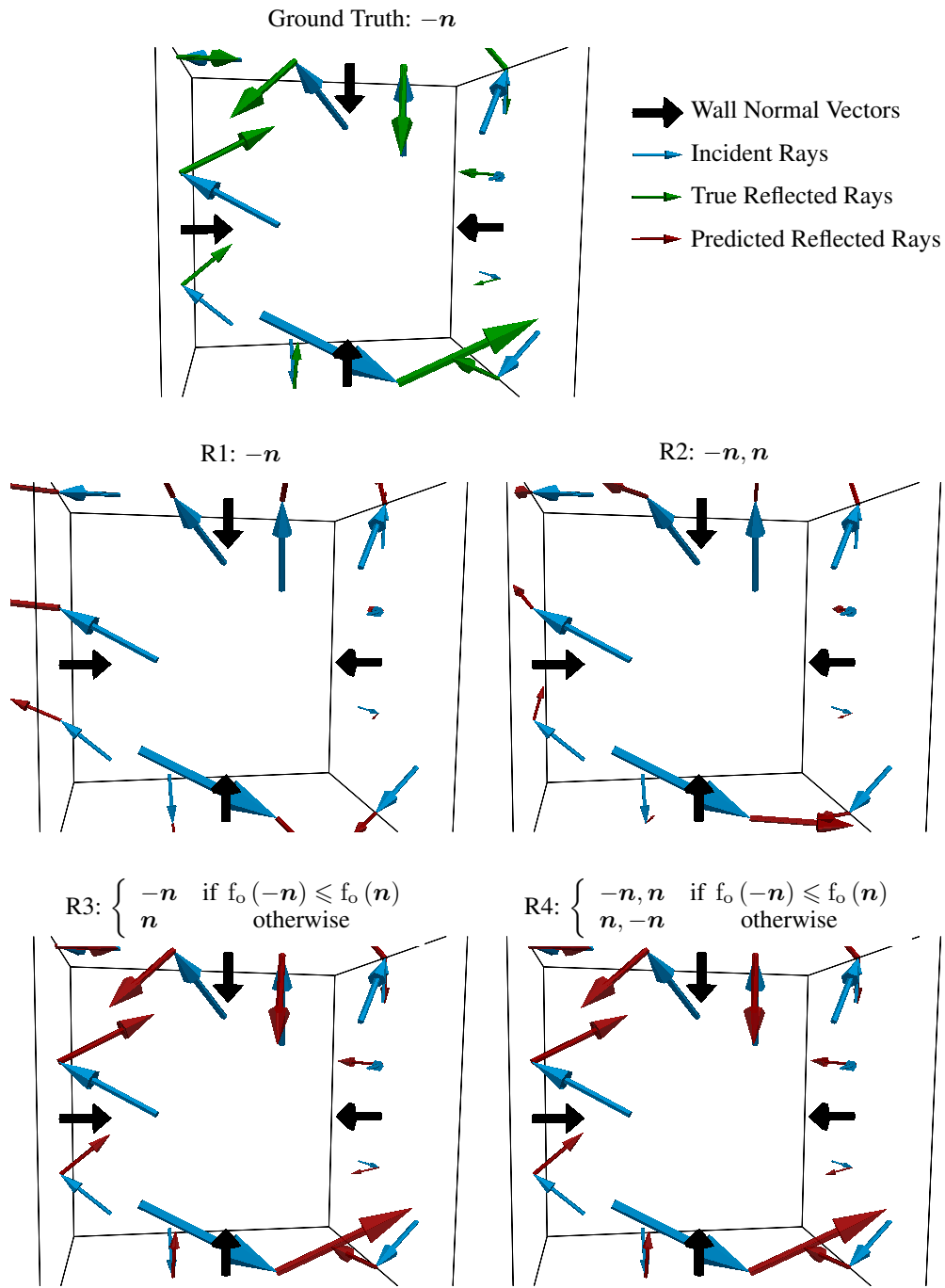

We use normal vectors of these wall, that point towards the outer of the cube. When evaluating the performance of the trained models, we observe decent predictions, if the orientation of the normal vectors describing the inclination of the walls was equal to the training data (see R1-R4 in Figure \ref{reflTrain}). However, for inverted normal vectors in the test set, only networks which are trained by taking a canonical orientation into account work well (see R3, R4 in Figure \ref{reflTrainInv}).

\clearpage

\subsection{Simulation Experiment}
\label{simExp}

We compare three versions for including normal vectors as boundary node features for the hopper particle flow:
\begin{itemize}
    \item not including normal vector information, filling six node features up with zero entries instead (V1)
    \item including single normal vector orientation, which is given by the triangle corner point order of the mesh (V2)
    \item including both normal vector orientations (six features) (V3)
\end{itemize}
From an information perspective, it should be noted that (i) distance information (scalar distance and relative distance vectors) to the walls is  present in the edge features of the graph and (ii) in most cases the used normal vectors were oriented towards the outside of relevant border walls.

The different particle distribution trajectories obtained by the three versions are compared by computing the Earth Movers distances~\citep[][EMD]{emd2011, flamary2017pot} between the trajectories from the machine learning model and the simulator. We use Euclidean distances for the cost matrix, which we compute at time steps $2^0, 2^1, \cdots, 2^{16}$ for 5 training trajectories and 5 test trajectories. Table \ref{tbl_nvec} shows the means ($\mu$) and standard deviations ($\sigma$) of EMD values at different time steps and from 5 different training and test trajectories. A paired Wilcoxon test on the concatenated trajectories, shows that V3 significantly outperforms V1 (p-value 2.42e-04) and V2 (p-value 1.50e-03) on the test data. 

Interestingly, there is less significance on the training data, which might indicate that the usage of orientation-independent features to represent walls, helps to improve generalization performance, while it might not be that helpful for optimization purposes alone.

\begin{table}[ht]
\caption{Usage of normal vector as node feature for the particle flow through a hopper. The table summarizes means ($\mu$) and standard deviations ($\sigma$) of the EMD for the different versions and shows the results of a paired Wilcoxon test.}
\label{tbl_nvec}
\begin{center}
\makebox[ \textwidth ]{
\begin{tabular}{ll|lll|lll}
\multicolumn{2}{c|}{\multirow{3}{*}{\bf \shortstack[c]{Version}}}
&\multicolumn{3}{c|}{\bf Train}
&\multicolumn{3}{c}{\bf Test}
\\
\multicolumn{2}{c|}{}
&\multicolumn{1}{c}{\multirow{2}{*}{\bf $\mu$}}  
&\multicolumn{1}{c}{\multirow{2}{*}{\bf $\sigma$}}
&\multicolumn{1}{c|}{\multirow{2}{*}{ \shortstack[c]{\small{p-value} \\ \tiny{Row < V3}}}}
&\multicolumn{1}{c}{\multirow{2}{*}{\bf $\mu$}}
&\multicolumn{1}{c}{\multirow{2}{*}{\bf $\sigma$}}
&\multicolumn{1}{c}{\multirow{2}{*}{ \shortstack[c]{\small{p-value} \\ \tiny{Row < V3}}}}
\\
\multicolumn{2}{c|}{}
&\multicolumn{3}{c|}{}
&\multicolumn{3}{c}{}
\\
\hline \multicolumn{2}{c|}{}
&\multicolumn{3}{c|}{}
&\multicolumn{3}{c}{}
\\

V1 & No normal vector     & 5.06e-05 & 1.17e-04 & 2.36e-02 & 6.80e-05 & 1.59e-04 & 2.42e-04\\
V2 & Single normal vector & 1.15e-04 & 3.84e-04 & 3.40e-03 & 1.21e-04 & 4.33e-04 & 1.50e-03\\
V3 & Both orientations    & 5.99e-05 & 1.77e-04 &          & 6.36e-05 & 2.06e-04 &         \\
\end{tabular}
}
\end{center}
\end{table}

\section{Mixing Entropy}
\label{mixent}

At time $t$, the local entropies $s(\Bx_{klm}, t)$ at grid cells, represented by a point $\Bx_{klm}$, are computed from particle counts $n_{+1}(\Bx_{klm}, t), n_{-1}(\Bx_{klm}, t)$ of the respective classes at the grid cells, where $n(\Bx_{klm}, t)=n_{+1}(\Bx_{klm}, t)+n_{-1}(\Bx_{klm}, t)$. The mixing entropy $\text{S}(t)$ is then computed by the formulas given by eqs \ref{eq_entropy}.

\begin{align}
\label{eq_entropy}
f_{\pm1}(\Bx_{klm}, t)&=\frac{n_{\pm1}(\Bx_{klm}, t)}{n_{+1}(\Bx_{klm}, t)+n_{-1}(\Bx_{klm}, t)}\\
s(\Bx_{klm}, t) &= - f_{+1}(\Bx_{klm}, t) \log f_{+1}(\Bx_{klm}, t) - f_{-1}(\Bx_{klm}, t) \log f_{-1}(\Bx_{klm}, t) \nonumber \\
\text{S}(t) &= \frac{1}{\sum\limits_{k,l,m} n(\Bx_{klm}, t)} \sum_{k,l,m} n(\Bx_{klm}, t) s(\Bx_{klm}, t) \nonumber
\end{align}

\clearpage


\end{document}